\documentclass[11pt]{article}

\usepackage{acl}

\usepackage{times}
\usepackage{latexsym}
\usepackage{booktabs}
\usepackage{tabularx}
\usepackage{graphicx}
\usepackage{latexsym}
\usepackage{CJKutf8}
\usepackage{calc}
\usepackage{amsmath} 
\usepackage{subcaption}
\usepackage{url}
\usepackage{amssymb}
\usepackage{enumitem}
\usepackage{graphicx}
\usepackage{subcaption}  
\usepackage{fontawesome5}
\usepackage{hyperref}

\renewcommand{\arraystretch}{1.3} 

\usepackage[T1]{fontenc}

\usepackage[utf8]{inputenc}

\usepackage{microtype}

\usepackage{inconsolata}

\usepackage{graphicx}

\title{Tracing Multilingual Knowledge Acquisition Dynamics in Domain Adaptation: A Case Study of English-Japanese Biomedical Adaptation}

\author{
  Xin Zhao\textsuperscript{\rm A,B},\quad
  Naoki Yoshinaga\textsuperscript{\rm C},\quad
  Yuma Tsuta\textsuperscript{\rm B}\thanks{Present address: Fixstars Corporation},\quad
  Akiko Aizawa\textsuperscript{\rm B} \\
  \textsuperscript{\rm A}The University of Tokyo \quad
  \textsuperscript{\rm B}National Institute of Informatics \\
  \textsuperscript{\rm C}Institute of Industrial Science, The University of Tokyo \\
  \texttt{xzhao@tkl.iis.u-tokyo.ac.jp, ynaga@iis.u-tokyo.ac.jp, aizawa@nii.ac.jp}
}

\begin{document}
\maketitle
\begin{abstract}
Multilingual domain adaptation (ML-DA) is widely used to learn new domain knowledge across languages into large language models (LLMs). 
Although many methods have been proposed to improve domain adaptation, the mechanisms of multilingual knowledge acquisition, how domain knowledge is learned within a language and transferred across languages, remain underexplored.
This gap leads to suboptimal performance, particularly in low-resource settings.
This work examines the learning dynamics of LLMs during ML-DA. 
Because prior ML-DA studies often train and evaluate on datasets with mismatched knowledge coverage, we propose \textbf{AdaXEval}, an adaptive evaluation method that builds multiple-choice QA datasets from the same bilingual domain corpus used for training, thereby directly studying multilingual knowledge acquisition.
Through continual training of LLMs with diverse data recipes, we track how LLMs acquire domain facts and pinpoint the mechanism behind the transformation process from domain training data to knowledge. 
Our experiments on a 13B English-Japanese bilingual LLM reveal that cross-lingual transfer remains challenging despite a high-quality bilingual corpus.
The code has been released.
\href{https://anonymous.4open.science/r/mlda-eval-2CD5}{\faGithub\ MLDA-Eval}

\end{abstract}

\section{Introduction}
\label{sec:intro}



Large language models (LLMs) trained on general-domain corpora perform well on diverse tasks but struggle in specialized domains~\cite{da-jang2022continualknowledgelearninglanguage, da-jang-etal-2022-temporalwiki}.
Domain adaptation addresses this by continually training LLMs on domain-specific data to enhance expertise~\cite{da-bio-jiang2024, da-bio-laïking2024, da-financial2024}.
Although prior work has explored strategies such as data augmentation and cross-lingual transfer to improve adaptation efficiency on low-resource settings~\cite{fang-etal-2023-chatgpt, gao-etal-2024-multilingual}, the mechanisms underlying effective domain knowledge acquisition and transfer remain insufficiently understood.

Understanding domain adaptation requires examining how LLMs acquire and internalize facts from domain data.
Prior studies show that factual knowledge accumulates gradually through repeated exposures, shaped by data frequency and model scale~\cite{chang2024largelanguagemodelsacquire, liu2025tracing, zhao-etal-2024-tracing}.
Moreover, multilingual settings introduce additional complexity, as equivalent facts may be encoded differently across languages~\cite{mondal-etal-2025-language, zhao-etal-2024-tracing}.
However, Most analyses focus on predefined relational probes rather than real domain facts with complex structures and specialized terminology.
Moreover, the link between training data and acquired knowledge also remains underexplored, which is crucial to optimizing domain adaptation strategies.

Our work aims to investigate the process of knowledge acquisition in domain adaptation from a mechanistic perspective. 
Specifically, we seek to understand, during the continual-training process on the domain corpus, how domain facts are \textit{memorized} and \textit{generalized} across different linguistic contexts, including both \textit{intralingual} (within a language) and \textit{interlingual} (across languages) variations, and to identify the key factors that facilitate effective knowledge acquisition and transfer.
To achieve the goal, we focus on three questions: 
\begin{description}[itemsep=0pt,topsep=1pt,parsep=1pt]
    \item[\textbf{RQ1: }] How to effectively evaluate the domain knowledge acquisition from diverse aspects?
    \item[\textbf{RQ2: }] What is the mechanism behind the transformation from training data to knowledge?
    \item[\textbf{RQ3: }] What factors are critical to achieve cross-lingual transfer?
\end{description}

Existing approaches to evaluate domain knowledge primarily rely on public benchmarks~\cite{eval-bio-singhal2022largelanguagemodelsencode,eval-jiang-etal-2025-jmedbench} or training loss analysis~\cite{zucchet2025languagemodelslearnfacts, liu2025tracing}. 
However, such benchmarks offer limited coverage for low-resource settings and fail to capture knowledge generalization abilities.
Moreover, the misalignment between training data and benchmark knowledge coverage makes evaluation an imperfect reflection of acquired knowledge.
To address these gaps and resolve \textbf{RQ1}, we propose \textbf{AdaXEval}, an adaptive domain knowledge evaluation data generation pipeline. 
AdaXEval automatically constructs multiple-choice datasets to evaluate knowledge \textit{memorization}, intralingual generalization (\textit{paraphrase}), and interlingual generalization (\textit{cross-lingual transfer}).
AdaXEval operates on either a monolingual or a bilingual domain corpus, where the latter is required for cross-lingual transfer evaluation, ensuring broad applicability across rare domains and low-resource languages.
Human annotation from multiple perspectives confirms that AdaXEval provides reliable evaluation.

We next investigate how training data is dynamically transformed into knowledge (\textbf{RQ2}).
Specifically, we conduct a case study of Japanese biomedicine domain adaptation using a 13B English/Japanese bilingual LLM~\cite{llmjp2024llmjpcrossorganizationalprojectresearch}, with English serving as a comparison and source for knowledge transfer.  
We begin with monolingual continual training on both English and Japanese using the J-STAGE corpus, which contains biomedical documents for both languages. 
By evaluating training checkpoints with AdaXEval-generated datasets, we observe a gradual knowledge acquisition process for cloze queries and paraphrases; however, LLM struggles to achieve cross-lingual transfer.
Further analysis reveals that knowledge is acquired as losses of correct options are shielded from rapid growth due to the model's overfitting to training data, which we term \textbf{loss shielding}. 
This acquisition eventually plateaus as training causes the model to overfit significantly to the training data, resulting in a substantial increase in loss across all options in evaluation instances. 
Examining losses on perturbed training data reveals that LLMs readily overfit to fixed token sequences in the training data, even under minor noise.

Finally, we investigate key factors influencing cross-lingual transfer of domain knowledge through \textbf{multilingual continual training} with diverse data recipes (\textbf{RQ3}).
We focus on translation and romanization strategies to enhance transfer.
Our results show that cross-lingual token overlap in related domains is crucial for effective knowledge transfer.
Nonetheless, even with high-quality alignment signals such as translations, cross-lingual transfer remains challenging, underscoring the need for more effective methods.


\section{Knowledge Acquisition Evaluation}
\label{sec:evaluation}
To effectively evaluate domain adaptation in low-resource scenarios, we propose AdaXEval, an adaptive pipeline for generating evaluation datasets.

\begin{figure*}[t]
    \centering
    \includegraphics[width=0.9\linewidth]{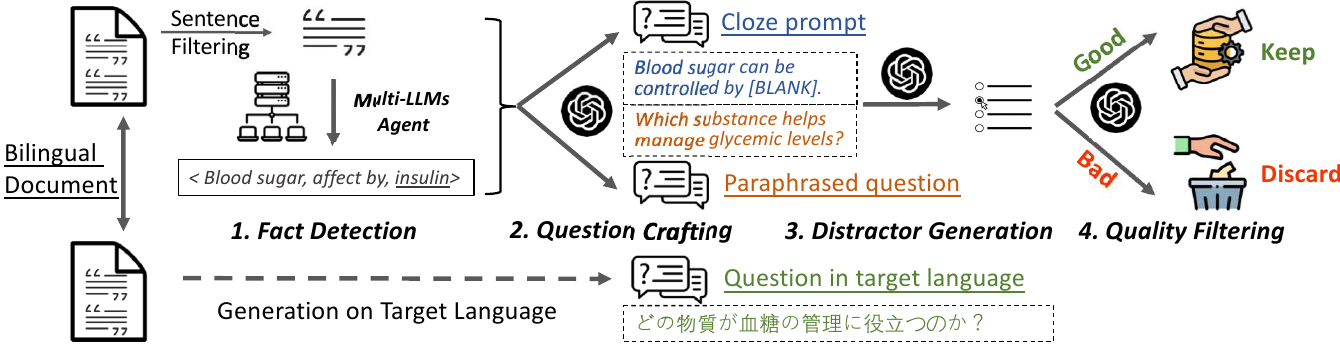}
    \caption{Overview of AdaXEval, a pipeline to adaptively generate domain knowledge evaluation datasets.}
    \label{fig:adaxeval}
\end{figure*}

\subsection{AdaXEval Pipeline}
AdaXEval is an adaptive evaluation pipeline that evaluates domain knowledge acquisition by generating evaluation datasets directly from the training corpus, ensuring evaluated facts stay aligned with the training data.
The pipeline includes four steps: fact detection, question crafting, distractor generation and quality filtering, as illustrated in Figure~\ref{fig:adaxeval}. 

\subsubsection{Fact Detection}
\label{subsec:fact-detect}
AdaXEval first detects sentences that may contain domain facts through a two-step strategy: named-entity-recognization (NER)-based sentence filtering and multi-agent fact detection.
First, domain-specific NER tools and linguistic heuristics are employed to identify sentences in the training corpora that contain multiple named entities.
Next, we design Chain-of-Thought (CoT) instructions to detect sentences containing domain facts from the filtered sentences, and extract triples in the format \texttt{$\langle$subject, relation, object$\rangle$} as the reference for question crafting.
Specifically, AdaXEval employs a multi-LLM agent for fact detection and triple extraction, estimating the overall confidence of the outputs, and adapting the top-confident extraction result to improve evaluation reliability.

\subsubsection{Queries Crafting}
\label{subsec:query-craft}
Given the factual sentence and referenced triple, we first prompt the LLM to generate reliable domain factual triples (\textit{e.g.}, \textit{$\langle$blood sugar level, can be controlled by, insulin$\rangle$}).
As complex domain knowledge cannot be easily formalized into named entities or predefined relations, subjects and objects are preferably named entities, though descriptive phrases are acceptable.
AdaXEval uses advanced LLMs to generate diverse question-answer pairs measuring three knowledge acquisition abilities.

\smallskip\noindent\textbf{1) Knowledge memorization} uses a cloze prompt with \texttt{[BLANK]} as the placeholder for the object (\textit{e.g.}, \textit{Blood sugar level can be controlled by [BLANK].}). Given the original sentence and refined triple as reference, we prompt the LLM to generate a cloze question that closely matches the original sentence to assess memorization exclusively.

\smallskip\noindent\textbf{2) Intralingual generalization}
assesses LLMs' ability to acquire knowledge using linguistic expressions that vary from those in the training corpus.
We design CoT instructions to let LLMs paraphrase the cloze queries into question-like style questions where different vocabulary is encouraged (\textit{e.g.}, \texttt{Which substance helps manage glycemic levels in the body?})

\smallskip\noindent\textbf{3) Interlingual generalization} measures how learned facts can be transferred across languages. 
While translation is a strong candidate, translating sentences that express domain knowledge is challenging due to the presence of specialized named entities, terminology, and concepts that lack direct equivalents across languages~\cite{liang-etal-2024-addressing}. 
To address this, we adapt AdaXEval to a bilingual domain corpus containing languages $X$ and $Y$, using the paraphrased dataset from language $X$ to evaluate cross-lingual transfer capabilities in $Y$.

\subsubsection{Distractor Generation} 
\label{subsec:distractor-generation}
AdaXEval then generates three plausible yet incorrect answer options that remain topically related but unambiguously wrong, while explicitly instructing the advanced 
LLM to avoid surface-level cues such as sequence length.

\subsubsection{Quality Filtering} 
Finally, AdaXEval uses the LLM to filter low-quality multiple-choice QA instances that fail to meet the requirements in \S~\ref{subsec:fact-detect},\ref{subsec:query-craft},\ref{subsec:distractor-generation}.


\subsection{Evaluation metric}
\label{para:eval_metric}

For each evaluation dataset, we follow \citet{eval-harness} to compute the average cross-entropy loss over the target tokens of possible answers and select the one with the highest generation possibility as the final answer. 
Specifically, for loss calculation of cloze queries, we use tokens before the \texttt{[BLANK]} as context and compute loss on the following tokens.
For paraphrases, we treat the question as context and measure only the loss of answer tokens.
We use prediction accuracy as the metric for knowledge acquisition. 
See Appendix~\ref{appendix:adaxeval-evaluation} for the mathematical formulation of the evaluation metric.

\subsection{Experimental Setup}
\label{subsec:adaxeval_exp}
\smallskip\noindent\textbf{Domain corpus:} 
Our study investigates biomedical domain adaptation in English–Japanese as a case study.
Specifically, we utilize the J-STAGE, an English-Japanese bilingual biomedical corpus (see \S~\ref{para:train_data}), as the data source for both model training and AdaXEval generation.

\smallskip\noindent\textbf{Details of Generation:}
We randomly sampled 10,000 bilingual documents to generate the evaluation dataset. 
We split abstracts into sentences and filter out sentences with fewer than two biomedical entities.  
For fact detection of filtered sentences, we use three open-source LLMs~\footnote{We employ open-source LLMs for local inference, as the large number of candidate sentences would otherwise incur substantial computational costs.} from different families to assess the confidence of whether the sentence is factual for each language. 
We then sum the confidence scores across the three LLMs and retain sentences with combined confidence scores greater than 2 (maximum 3).
Finally, we use GPT-4.1 to generate cloze queries, paraphrases, and three distractors for each instance. 
See Appendices~\ref{appendix:adaxeval_setting} and \ref{appendix:adaxeval_examples} for details of the generation process and statistical report of generated datasets.

\smallskip\noindent\textbf{Human Evaluation:}
To assess the quality of our generated datasets, we conduct a comprehensive human evaluation across four key components of the knowledge extraction and question generation pipeline, including triple extraction quality evaluation, cloze prompt evaluation, paraphrased question evaluation, and distractor quality evaluation. 
Overall, the evaluation result indicates that AdaXEval is able to generate high-quality evaluation data, meeting the requirements for assessing diverse knowledge acquisition abilities.
See Appendix~\ref{appendix:adaxeval_examples} for evaluation results and dataset examples, and Appendix~\ref{appendix:human_eval_guide} for the human evaluation guideline.

\section{Tracing Knowledge Acquisition}
\label{sec:tracing_knowledge}

This section examines the training dynamics of domain adaptation and explores the mechanism underlying the transformation from training data to knowledge in the monolingual setting. 
We perform monolingual continual training on English and Japanese using a bilingual biomedical dataset.

\subsection{Experimental Setup}

\smallskip\noindent\textbf{Data preparation:}\label{para:train_data}
We utilize a subset of the J-STAGE corpus, which comprises Japanese research papers with some abstracts translated into English.\footnote{Access to the dataset is restricted by the J-STAGE license, so it cannot be publicly released.}
Specifically, we select 614,444 Japanese and 404,643 English biomedical documents, paired one-to-one. 
These bilingual pairs provide source data for AdaXEval generation.
To strengthen domain adaptation and enable fine-grained analysis, we apply instruction pretraining as a data augmentation baseline~\cite{instr-pretrain-qa}. 
Biomedical instructions are generated from raw text using both rule-based mining patterns~\cite{instr-pretrain-rule} and LLM-based question-answer generation~\cite{instr-pretrain-qa}.
The raw documents are then combined with the generated instructions for continual training.
Details of the training dataset construction are provided in Appendix~\ref{appendix:data_generation}.

\smallskip\noindent\textbf{Training setup:}
We adopt llm-jp-3-13B~\cite{llmjp2024llmjpcrossorganizationalprojectresearch}, a strong Japanese–English bilingual LLM, as the base model for pretraining, owing to its superior language ability in both languages, particularly Japanese.
For each language, we cut off 0.5B tokens from the constructed corpus and train the data on llm-jp-3-13B for four epochs.
Training hyperparameters are detailed in Appendix~\ref{appendix:training_parameter}.

\begin{figure}[t]
  \centering
  \includegraphics[width=\linewidth]{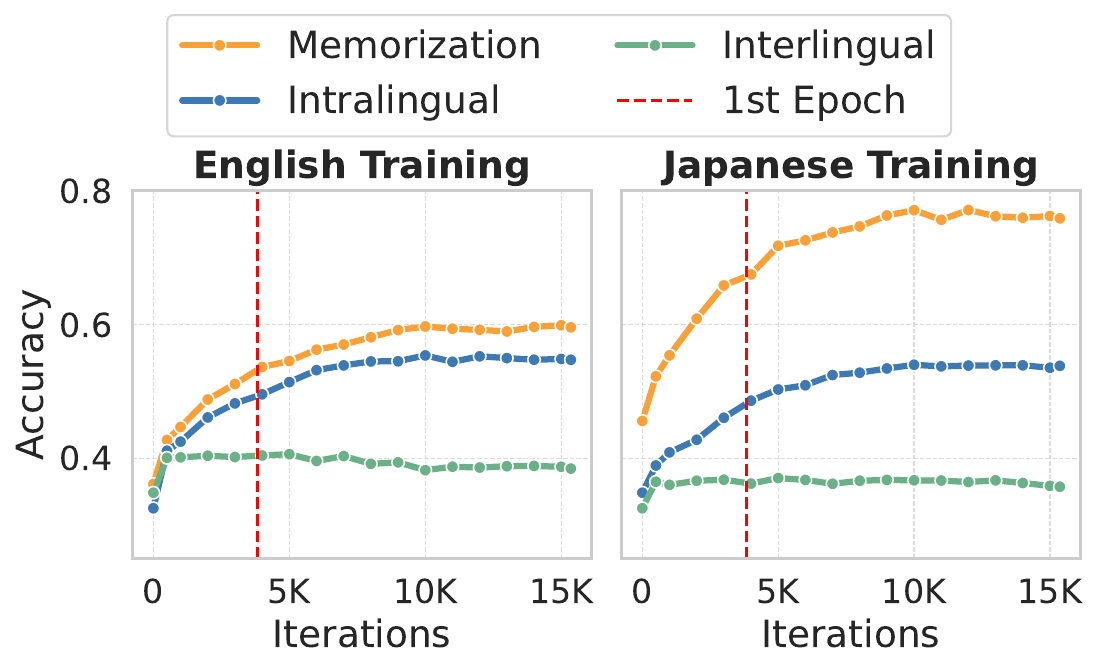}
  \caption{Dynamic knowledge acquisition evaluation after monolingual pretraining.}
  \label{fig:adaxeval-eval}
\end{figure}

\subsection{Tracing Performance Dynamics}
Figure~\ref{fig:adaxeval-eval} reports AdaXEval results for English and Japanese monolingual training.
The results indicate that domain knowledge is gradually acquired during training in both languages.

\smallskip\noindent\textbf{Memorization evaluation:}
Accuracy increases from 36.1\% to 59.6\% (\textbf{+23.5\%}) in English and from 45.7\% to 75.9\% (\textbf{+30.2\%}) in Japanese.
The higher post-training accuracy in Japanese partly reflects the stronger medical knowledge base of llm-jp-3-13B.
However, since the factual instances used for evaluation differ between languages, direct cross-lingual comparison is not strictly fair.

\smallskip\noindent\textbf{Intralingual generalization:}
Both languages exhibit strong performance on the paraphrased datasets, with accuracy increasing from 32.6\% to 54.7\% (\textbf{+22.1\%}) in English and from 34.9\% to 53.8\% (\textbf{+18.9\%}) in Japanese. 
Notably, the improvement in English paraphrases parallels the memorization gain, whereas it is about 10\% lower in Japanese, suggesting that the difficulty of intralingual generalization differs across languages.

\smallskip\noindent\textbf{Interlingual generalization:}
Figure~\ref{fig:adaxeval-eval} reveals that monolingual training results in limited cross-lingual knowledge transfer, yielding only \textbf{3.6\%} improvement in English-to-Japanese and \textbf{3.1\%} improvement in Japanese-to-English transfer.

\subsection{Knowledge Acquisition via Loss Shielding}

To elucidate the mechanisms of knowledge acquisition, this section analyzes the sequence loss of evaluation data to understand how knowledge is acquired during training.
We analyze loss as it directly drives predictions on our multiple-choice datasets (see \S~\ref{para:eval_metric}) and reflects the model’s generation behavior, where sequences with lower loss are more likely to be generated.

\begin{figure}[t]
  \centering
  \begin{subfigure}[t]{\linewidth}
    \centering
    \includegraphics[width=0.95\linewidth]{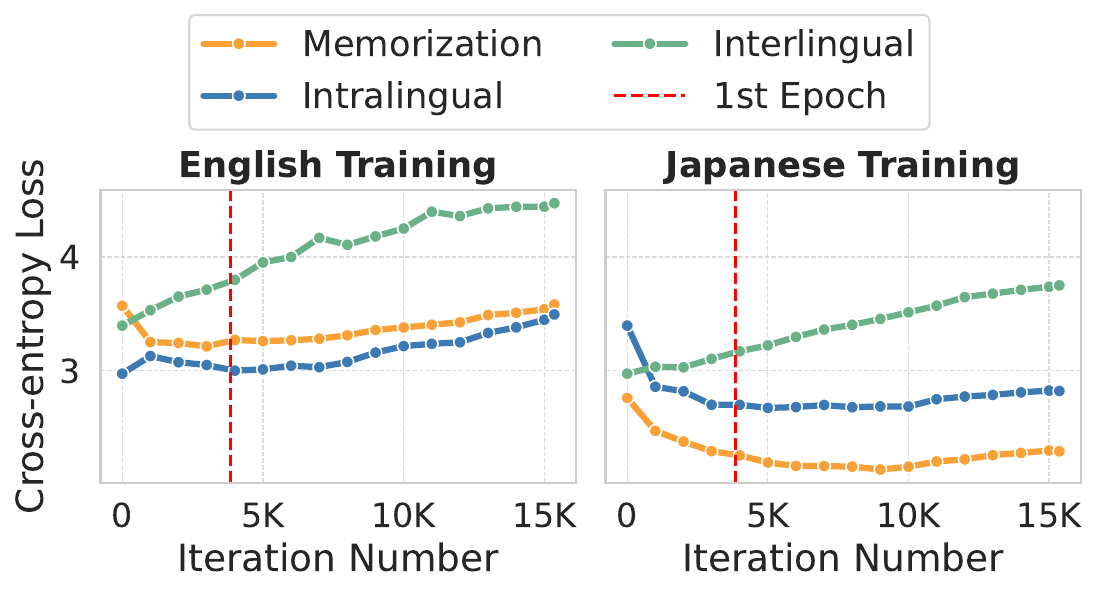}
    \caption{\textbf{Loss dynamics} of correct query–answer sequences.}
    \label{fig:adaxeval-loss}
  \end{subfigure}
  \vspace{0.6em} 
  \begin{subfigure}[t]{\linewidth}
    \centering
    \includegraphics[width=\linewidth]{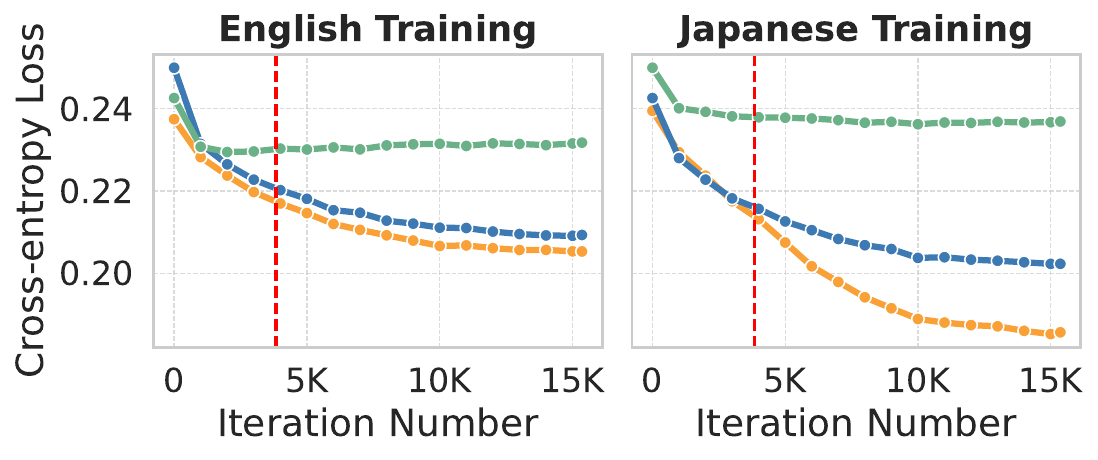}
    \caption{\textbf{Loss ratio dynamics} for sequences with correct answers relative to all candidates.}
    \label{fig:adaxeval-lossratio}
  \end{subfigure}
  \caption{Loss dynamics of datasets generated by AdaXEval during monolingual training.}
\end{figure}

\smallskip\noindent\textbf{(1) Training overfits to data, but the \emph{loss shielding} drives knowledge memorization.}
We calculate the sequence loss of queries paired with correct answers in the evaluation dataset obtained by AdaXEval.
Figure~\ref{fig:adaxeval-loss} shows the loss trajectory across training checkpoints for English and Japanese training.
On the cloze prompt dataset, the loss decreases in early training but rises in later training, suggesting that training causes the model to overfit to the training corpus.
However, Figure~\ref{fig:adaxeval-eval} reveals that memorization accuracy continues to improve until the 10K-th iteration.
To investigate this divergence, we measure the ratio of the correct-sequence loss to the total loss across all four options.
As shown in Figure~\ref{fig:adaxeval-lossratio}, this ratio mirrors the accuracy trend, suggesting that knowledge can still be memorized even under overfitting, since correct sequences are shielded from rapid loss growth, a phenomenon we term \textit{loss shielding}.
Figure~\ref{fig:adaxeval-lossratio} also explains the gap between cloze prompts and paraphrases, which the significant loss ratio gap in Japanese predicts, as shown in Figure~\ref{fig:adaxeval-eval}. 

\begin{figure}[t]
  \centering
  \includegraphics[width=\linewidth]{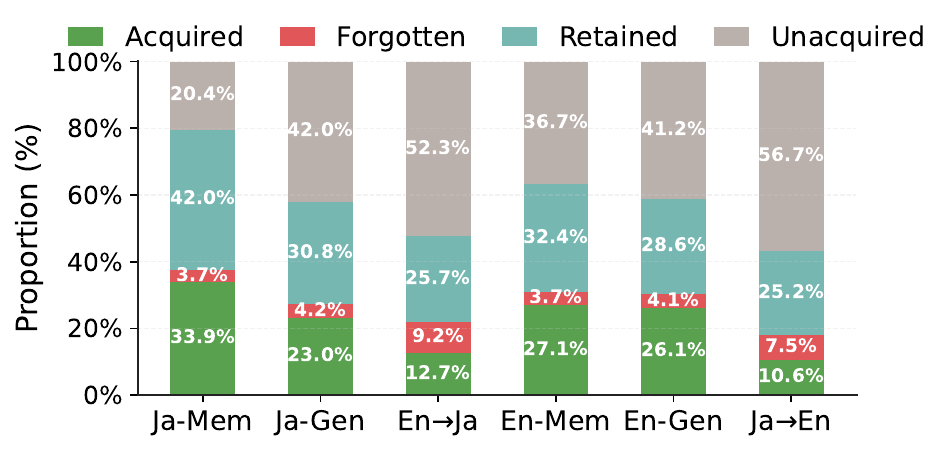}
  \caption{Instance state transitions before/after training.}
  \label{fig:knowledge-transition}
\end{figure}

\smallskip\noindent\textbf{(2) A trade-off exists between knowledge acquisition and forgetting.} 
For each instance, we check its state transition before and after training by examining the loss.
Figure~\ref{fig:knowledge-transition} shows the proportions of instances retained, acquired, forgotten, or unacquired during training.
Forgetting remains limited in monolingual evaluations, including both memorization and intralingual generalization.
In contrast, cross-lingual transfer exhibits a notable increase in forgotten cases, offsetting gains from newly acquired knowledge.
This suggests that while training in one language can introduce transferable knowledge, it also causes a decline in performance in other languages due to forgetting.

\begin{figure}[t]
  \centering
  \includegraphics[width=\linewidth]{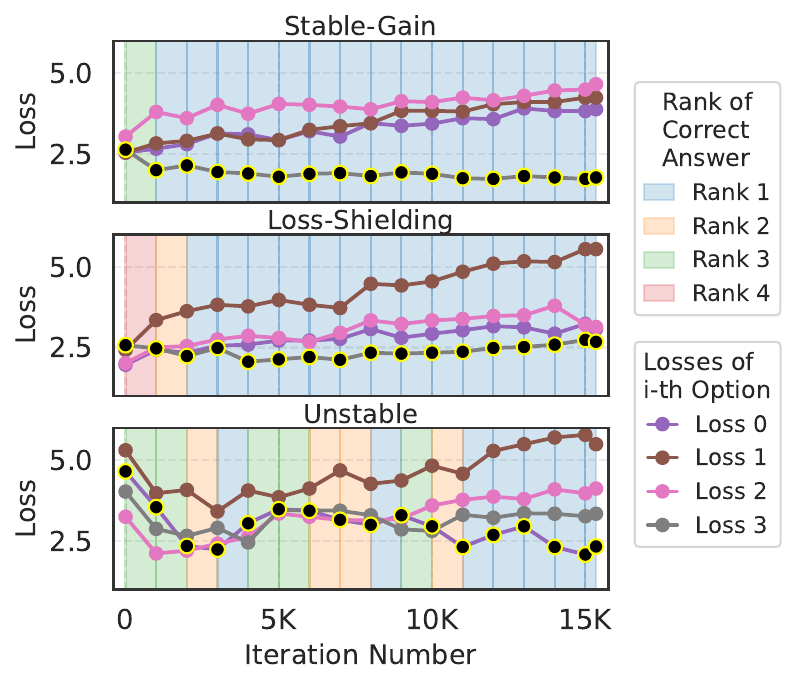}
  \caption{Acquired instances with different loss shapes. The line with bright circles indicates the correct answer.}
  \label{fig:instance-check-loss}
\end{figure}

\smallskip\noindent\textbf{(3) Instance-level case studies show diverse knowledge acquisition patterns.} 
We then examine the loss dynamics of instances acquired after training, analyzing all four options. 
We observe that the loss dynamics of correct answers follow distinct patterns: they either decrease steadily (Stable-Gain), increase while remaining lower than incorrect options (Loss-Shielding), or exhibit unstable behavior. 
Figure~\ref{fig:instance-check-loss} illustrates three examples corresponding to the three loss change patterns.

\section{From Training Data to Knowledge}
This section investigates how knowledge of unseen queries is derived from the training data.
Previous analyses have shown that the model tends to increase the loss for sequences that express the same domain knowledge as the training data but in different linguistic forms.
Although loss shielding allows the model to continue acquiring knowledge even as the loss for such sequences rises, this effect eventually fades once the loss grows too large.
Understanding what sustains and what breaks this shielding effect is therefore essential for developing more robust training strategies.
To this end, we introduce controlled perturbations into the training data by injecting noise under different rules, and track how their loss evolves during training.
Specifically, we randomly sample 2,000 sequences from Japanese monolingual training data and apply perturbations at both the token and sequence levels.

\subsection{Token-level Perturbation}
Token sequences are perturbed after tokenization using the methods described below. 

\begin{itemize}[leftmargin=1em,itemsep=0pt,topsep=1pt,parsep=0pt]
    \item \textbf{mask-X}: Replace X\% tokens with \texttt{<unk>} token.
    \item \textbf{random-X}: Replace X\% tokens with randomly sampled tokens from the tokenizer vocabulary.
    \item \textbf{delete-X}: Delete X\% tokens.
    \item \textbf{reorder-X@Y}: Reorder tokens to achieve an edit distance equal to X\% of the sequence length, with swaps restricted to a window of size Y.
    \item \textbf{monosyn-X}: Replace with Japanese synonyms.
    \item \textbf{mltlsyn-X}: Replace with English synonyms.
\end{itemize}

\begin{figure}[t]
  \centering
    \centering
    \includegraphics[width=0.95\linewidth]{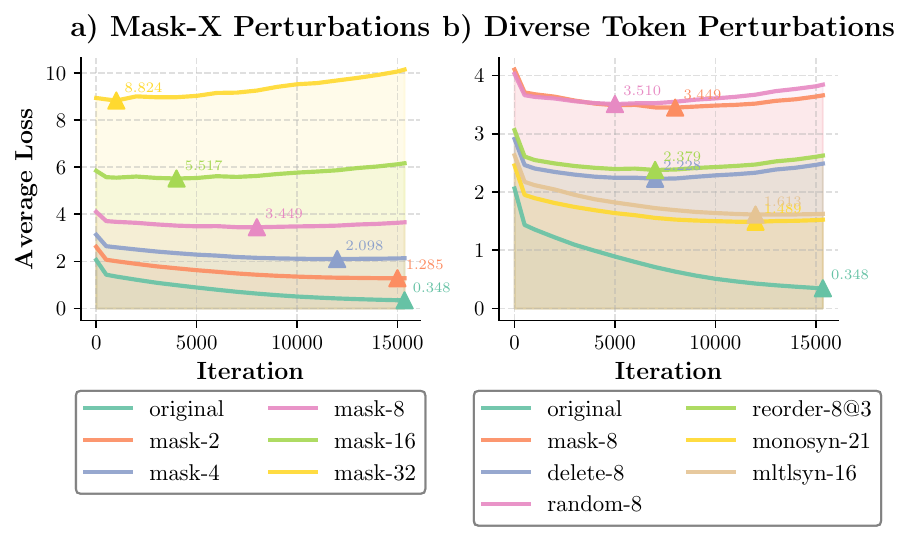}
    \caption{Loss dynamics of token-level perturbation.}
    \label{fig:token-perturbation}
\end{figure}

Specifically, we set \mbox{$X = 2^{1,\ldots,5}$} and \mbox{$Y = 2^{0,\ldots,4}$}. 
Synonym substitutions use the Japanese WordNet 2.0~\cite{wordnet-jp} to retrieve Japanese or English synonyms. 
Additional implementation details are provided in Appendix~\ref{appendix:sequence_perturb}.

\smallskip\noindent\textbf{(1) More token edits cause larger loss harm.}
Taking \emph{mask-X} as an example, Figure~\ref{fig:token-perturbation}(a) illustrates the loss dynamics across edited sequences compared to their original counterparts.
Triangles ($\blacktriangle$) mark the \emph{overfitting onset}, the point where minimum loss is attained before overfitting induces an upward trend.
The results indicate that while the loss of original sequences decreases steadily, masking tokens introduce a substantial initial loss increase and accelerate the onset of overfitting.
The analysis of other perturbation patterns leads to the same conclusion. 
See Appendix~\ref{appendix:sequence_perturb} for more results.

\smallskip\noindent\textbf{(2) Loss sensitivity varies with vocabulary and structural perturbations.}\label{para:token-vocab-insensitive}
Figure~\ref{fig:token-perturbation}(b) illustrates the loss variations across sequences with 8\% of tokens perturbed using diverse methods. 
Semantically aligned modifications (\emph{monosyn} and \emph{mltlsyn}), exhibit the least impact on loss, followed by structural alterations (\emph{reorder}, \emph{delete}) that avoid introducing new vocabulary. 
Perturbations introducing irrelevant tokens (\emph{mask}, \emph{random}) inflict the greatest harm, significantly increasing the initial loss and accelerating the onset of overfitting. 



\subsection{Sentence-level Perturbation}

We further perturb target sequences at the sentence level to simulate more realistic noise patterns, considering the following perturbation strategies.

\begin{itemize}[leftmargin=1em,itemsep=0pt,topsep=1pt,parsep=0pt]
    \item \textbf{partial-a}: Split each training document into four segments, then select the $a$-th segment.
    \item \textbf{syntax-X}: Rewrite $X$\% sentences, modifying only syntax without changing vocabulary.
    \item \textbf{lexicon-X}: Rewrite $X$\% sentences, modifying only vocabulary without changing syntax.
    \item \textbf{semantic-X}: Rewrite $X$\% sentences, allowing both syntactic and lexical changes.
    \item \textbf{translation-X}: Translate $X$\% sentences.
\end{itemize}

\begin{figure}[t]
  \centering
    \centering
    \includegraphics[width=0.95\linewidth]{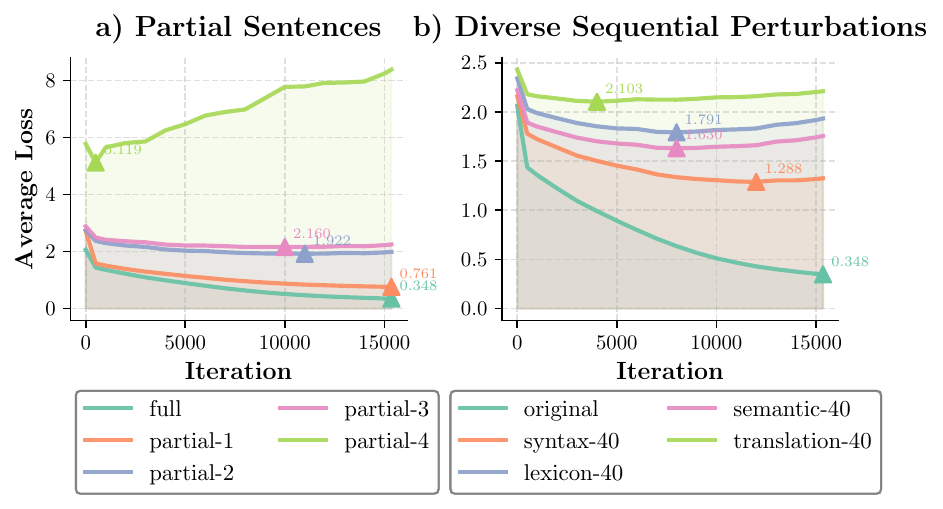}
    \caption{Loss dynamics of sequence-level perturbation.}
    \label{fig:sequence-perturbation}
\end{figure}

\smallskip\noindent\textbf{(1) Losses exhibit stronger dependence on prior context.}
Using partial sentences following the \emph{partial-a} strategy, Figure~\ref{fig:sequence-perturbation}(a) shows that sentences appearing later in a training document incur higher loss when evaluated in isolation. 
This suggests that losses depend heavily on prior context, which constrains knowledge acquisition and highlights the need for a more balanced, context-independent learning paradigm.

\smallskip\noindent\textbf{(2) Preserving vocabulary improves loss robustness.}  
Figure~\ref{fig:sequence-perturbation}(b) presents the loss dynamics for all rewriting patterns applied to 40\% of sentences, including both paraphrasing and translation. 
Among these patterns, syntax rewriting shows the most substantial loss shielding effect, followed by semantic rewriting. 
This is because semantic paraphrasing does not require extensive vocabulary replacement, resulting in fewer word substitutions compared to the lexicon paraphrases.
These results highlight the necessity of incorporating lexicon-focused paraphrasing during training to improve models’ ability to generalize knowledge across diverse test inputs, consistent with the observations in \S~\ref{para:token-vocab-insensitive}.



\section{Bridging Languages in Domain Adaptation}

\subsection{What Helps Cross-lingual Transfer?}
\begin{figure}[t]
  \centering
  \includegraphics[width=0.9\linewidth]{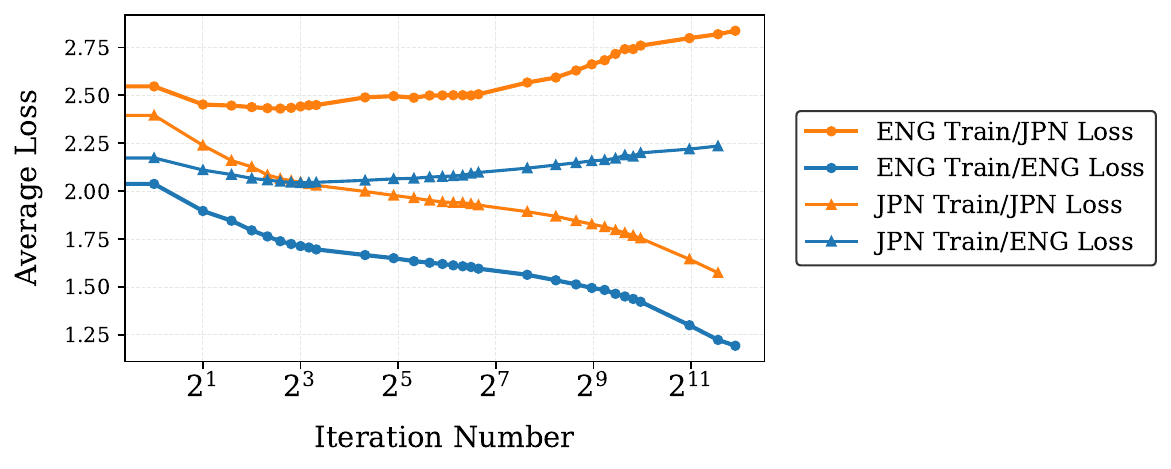}
  \caption{Losses on English/Japanese training data.}
  \label{fig:seq-loss}
\end{figure}

We examine how cross-lingual transfer occurs by tracking the loss of target-language documents in models trained on source-language data.
As the AdaXEval interlingual loss continues to increase (Figure~\ref{fig:adaxeval-loss}), tracing loss on evaluation data becomes unreliable.
Instead, we examine the loss of the training documents across languages, serving as indirect evidence.
Specifically, we sample 1,000 documents for each language and measure per-token loss on both datasets under two monolingual training settings.
Since cross-lingual transfer converges rapidly within the first 1,000 iterations, we analyze the loss dynamics at a finer granularity during the first epoch. 
The results are shown in Figure~\ref{fig:seq-loss}.
We observe that in-language loss consistently decreases, whereas cross-language loss decreases only during the initial iterations, reflecting the rapid but short-lived interlingual generalization.

\begin{figure}[t]
  \centering
  \includegraphics[width=0.9\linewidth]{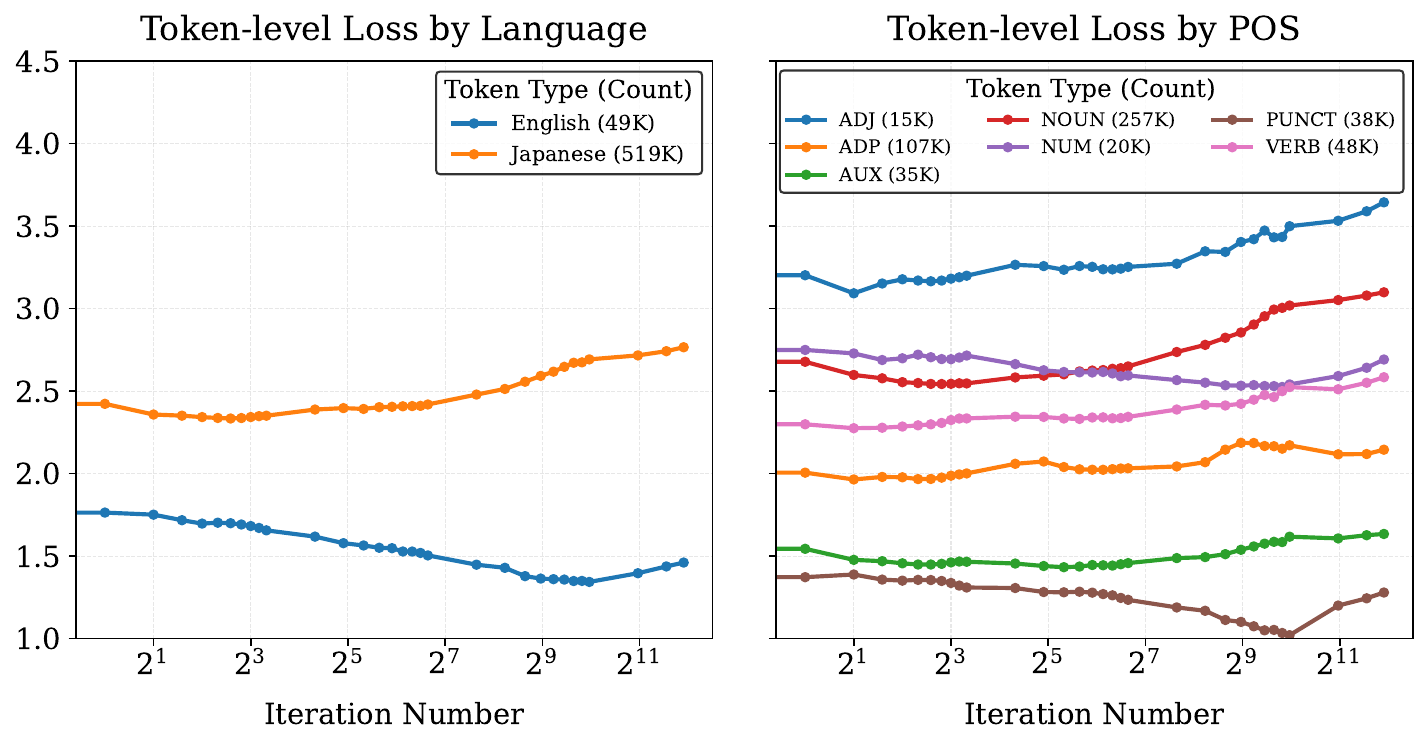}
  \caption{Japanese per-token loss by English training.}
  \label{fig:token-loss}
\end{figure}

To investigate what tokens contribute to the initial loss reduction, we further measure the per-token cross-entropy loss within the given sequences.
Specifically, we examine the impact of two linguistic characteristics on loss changes: language and part-of-speech (POS).
For language, tokens are classified by the language in which they occur, either Japanese or English; For POS, tokens are grouped by their syntactic category.
Figure~\ref{fig:token-loss} shows token-level loss dynamics on Japanese documents during English training. 
Language-based token analysis indicates that English tokens embedded in Japanese documents benefit from English training. 
POS-based analysis supports the findings, as only numerical (NUM) and punctuation (PUNCT) tokens, both present in the English corpus, exhibit notable reductions in loss.
This finding suggests that tokens occurring in the training corpus experience greater learning gains. 

\subsection{Cross-lingual Transfer Enhancement}
In this section, we explore the factors in the training corpus that can drive improvements in cross-lingual transfer during domain adaptation.
Due to the limited cross-lingual transfer ability in monolingual training, we switch to multilingual domain adaptation.
Specifically, we investigate: what types of multilingual corpora can facilitate knowledge transfer during domain adaptation?

\subsubsection{Multilingual Continual Training}
\label{subsec:multi-train-exp}

To examine the factors that facilitate cross-lingual transfer, we construct a series of multilingual corpora and evaluate their effectiveness.
To ensure a fair comparison, each training corpus is composed of two parts: a \textbf{knowledge injection corpus} $\mathcal{C_K}$ and a \textbf{cross-lingual transfer enhancement corpus $\mathcal{C_T}$}. 
The $\mathcal{C_K}$ contains the target knowledge expressed exclusively in the source language $X$, while $\mathcal{C_T}$, by contrast, does not provide any new knowledge in the target domain; instead, it serves only to establish linguistic connections between source language $X$ and target language $Y$.
Finally, we evaluate cross-lingual knowledge transfer by measuring the model’s ability to acquire knowledge in language $Y$, using the same evaluation metrics introduced in \S~\ref{para:eval_metric}. 
We fix the source language $X$ as English and the target language $Y$ as Japanese to facilitate analysis in this section. 
See Appendix~\ref{appendix:cross-lingual-transfer-evaluation} for the evaluation on reverse transfer direction. 

\smallskip\noindent\textbf{Cross-lingual transfer enhancement corpora:}  
We construct diverse corpora to enhance cross-lingual transfer using two primary strategies: \emph{translation} and \emph{romanization}. 
Both can build token-level connections between English and Japanese, where romanization requires significantly less effort for data collection. 
As baselines, we consider an empty cross-lingual corpus (\(\mathcal{C_T} = \emptyset\)) (\textbf{Monolingual}) and a strong domain-specific baseline using J-STAGE Japanese data, with documents related to the AdaXEval evaluation filtered out (\textbf{Medical-Japanese}).

To examine which translation data are most effective for domain adaptation, we prepare three types of bilingual corpora and use them to generate translation instructions as $\mathcal{C_T}$:  
\begin{itemize}[leftmargin=1em,itemsep=1pt,topsep=2pt,parsep=1pt]
    \item \textbf{JParaCrawl (Balanced-Translation)}: An English–Japanese web-crawled corpus covering diverse domains~\cite{jparacrawl}.  
    \item \textbf{ASPEC (Science-Translation)}: A multilingual corpus containing academic paper abstracts across various scientific fields~\cite{ASPEC}. Documents in the medical and chemical domains are excluded to distinguish them from the target medical domain.  
    \item \textbf{J-STAGE (Medical-Translation)}: J-STAGE represents the closest domain to our target. 
    We filter out all documents included in the AdaXEval evaluation datasets to avoid contamination.  
\end{itemize}

To evaluate the romanization strategy, we construct a medical romanization dataset (\textbf{Medical-Roman}).  
We convert J-STAGE Japanese text to romaji using \texttt{cutlet}\footnote{\url{https://github.com/polm/cutlet}}, an open-source tool for romanization.  
We then generate romanization instructions to link the Latin script of English with the Japanese script (kanji, hiragana, etc.).  
Finally, we create translation instructions between romanized Japanese and English based on J-STAGE (\textbf{Medical-Roman2En}), which serves as a comparison group.

\smallskip\noindent\textbf{Details of training:} 
Focusing on knowledge transfer from English to Japanese, we prepare two corpora, $\mathcal{C}_K$ and $\mathcal{C}_T$, each containing 0.5 billion tokens, except for the \textbf{Monolingual} baseline.  
The English J-STAGE pretraining data serve as $\mathcal{C}_K$.  
For $\mathcal{C}_T$, we select seven candidate datasets as described in Sec.~\ref{subsec:multi-train-exp}.  
We then combine the two 0.5B-tokens corpora into a single 1B-token corpus, shuffle it, and train llm-jp-3-13B with it for one epoch.
See Appendix~\ref{appendix:training_settings} for more training details.

\subsubsection{Cross-lingual Transfer Evaluation}
\label{subsec:cross-lingual-transfer-evaluation}

\begin{figure}[t]
  \centering
  \includegraphics[width=\linewidth]{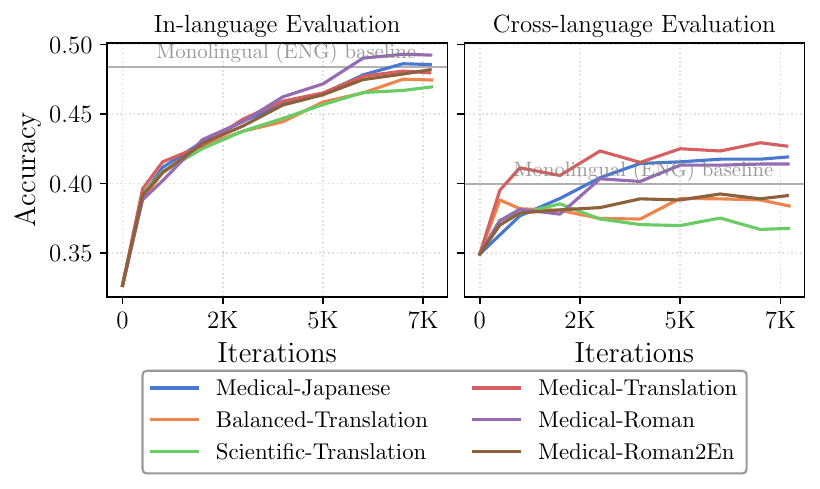}
  \caption{Enhancing cross-lingual transfer using diverse multilingual domain corpora.}
  \label{fig:crosslingual-transfer-enhancement}
\end{figure}

\begin{figure}[t]
  \centering
  \includegraphics[width=\linewidth]{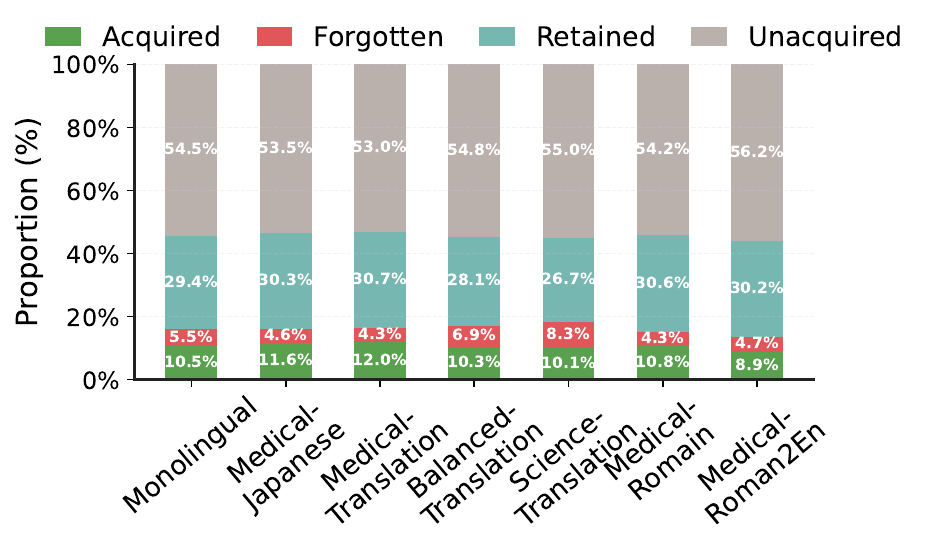}
  \caption{State transitions by cross-lingual transfer.}
  \label{fig:multilingual-state-transition}
\end{figure}

\smallskip\noindent\textbf{(1) Domain-specific corpus enhances cross-lingual transfer.} Figure~\ref{fig:crosslingual-transfer-enhancement} presents the accuracy dynamics of multilingual domain adaptation, comparing two baselines with three translation datasets and two romanization datasets.
Among these, only the corpus constructed from domain-specific data (\textbf{Medical-Japanese/Roman/Translation}) surpasses the \textbf{Monolingual} baseline, and only \textbf{Medical-Translation} achieves higher performance than both baselines.
These results indicate that effective cross-lingual transfer of domain knowledge requires domain-specific signals; general cross-lingual enhancement methods fail to yield comparable improvements for domain knowledge.
Figure~\ref{fig:multilingual-state-transition} further illustrates the state transitions induced by cross-lingual transfer, showing that corpora yielding performance gains both increase acquired instances and reduce forgotten ones, whereas the other datasets perform worse on both aspects.
Finally, the stable performance across all recipes in the in-language evaluation indicates that using additional corpora unrelated to the target knowledge does not impair in-language knowledge acquisition.



\smallskip\noindent\textbf{(2) Achieving effective cross-lingual transfer of domain knowledge is challenging.}  
Although both \textbf{Medical-Translation} yield improvements, the gains are still limited (around 3\%) relative to the doubled training cost and additional dataset construction effort. 
This highlights the need for developing more efficient methods for cross-lingual domain knowledge transfer.

\section{Related Work}

\subsection{Domain Knowledge Evaluation}
Domain knowledge is typically evaluated using public benchmarks~\cite{eval-xie2024finbenholisticfinancialbenchmark, eval-jiang-etal-2025-jmedbench}. However, such benchmarks are often unavailable for low-resource languages or specialized domains. 
General benchmarks also cannot fully assess intralingual/interlingual generalization, which requires paraphrase or translation datasets.
Moreover, the misalignment between training data and benchmark knowledge coverage makes evaluation an imperfect reflection of acquired knowledge.
Conversely, some studies use loss reduction as evidence of knowledge acquisition~\cite{chang2024largelanguagemodelsacquire, zucchet2025languagemodelslearnfacts, liu2025tracing}, but loss does not necessarily reflect domain-specific knowledge that users care about.

\subsection{Mechanisms of Knowledge Acquisition}
LLMs have been widely studied for their capacity to store and retrieve factual knowledge~\cite{petroni2019language, hernandez2023measuring, chang2024largelanguagemodelsacquire, zhao-etal-2024-matters}, spurring interest in understanding how such knowledge is encoded and accessed~\cite{dai-etal-2022-knowledge, wang-etal-2022-finding-skill, niu2024what, zhao-etal-2025-neuron}.
Recent work has investigated knowledge acquisition dynamics by analyzing intermediate checkpoints during training.
\citet{chang2024largelanguagemodelsacquire} shows that LLMs accumulate factual knowledge through repeated exposures, gradually increasing recall likelihood with each encounter.
Key factors influencing acquisition include fact frequency, model scale, and batch size~\cite{liu2025tracing, zhao-etal-2024-tracing}.

Meanwhile, cross-lingual transfer ability has also garnered considerable attention for enabling efficient knowledge acquisition across languages.
However, transferring factual knowledge across languages remains notably challenging~\cite{liu2025tracing, zhao-etal-2024-tracing}.
Facts expressed in different languages may be stored as distinct representations~\cite{chen2023journeycenterknowledgeneurons, zhao-etal-2024-tracing, mondal-etal-2025-language}, and only certain relational types transfer effectively~\cite{liu2025tracing, zhao-etal-2024-tracing}.
Despite these advances, existing studies focus on predefined relational facts~\cite{zhao-etal-2024-tracing, liu2025tracing}, which inadequately represent complex knowledge in specialized domains. Moreover, the connection between training data and acquired knowledge remains underexplored, which is critical for designing optimal training recipes.

\subsection{Cross-lingual Knowledge Transfer}
Cross-lingual transfer in domain adaptation has been studied through various strategies that aim to bridge language gaps in knowledge transfer.  
Translation data are widely used in training LLMs to enhance cross-lingual knowledge transfer~\cite{multilingual-recipe,multilingual-general-1,multilingual-knowledge-1}, supporting applications such as machine translation~\cite{multilingual-general-1} and instruction following~\cite{multilingual-instruction-1,multilingual-instruction-2}.  
However, since in-domain translation data are often scarce, it remains unclear whether translation can effectively transfer complex and sparse domain knowledge across languages.  
Furthermore, the romanization strategy reduces script barriers by converting text into romanized forms, thereby aligning linguistic representations with English~\cite{multilingual-romanization-1,multilingual-romanization-2,multilingual-romanization-3}.  
Finally, code-switching has been shown to be effective in aligning token semantics across languages by interleaving tokens from multiple languages within the same context~\cite{multilingual-codeswitch-2, multilingual-codeswitch-1}.
However, it is still uncertain whether these alignment signals are sufficient for transferring complex domain knowledge across languages.

\section{Conclusions}

In this paper, we studied how LLMs acquire domain knowledge and transfer it across languages.  
We proposed \textbf{AdaXEval}, an adaptive evaluation pipeline that automatically generates datasets to assess domain knowledge across memorization, intralingual generalization, and cross-lingual transfer.  
Using AdaXEval, we conduct a case study focusing on English-Japanese biomedical domain adaptation. 
We analyze the training dynamics of domain adaptation and find that knowledge acquisition is driven by \textbf{loss shielding}, wherein overfitting raises losses on irrelevant representations more rapidly than on relevant ones.  
Through \textbf{perturbation analysis}, we further reveal the sensitivity of LLMs to training data, offering practical guidance for developing more robust training paradigms.
We also identified key factors that facilitate cross-lingual transfer through \textbf{controlled multilingual continual training}, showing that the presence of cross-lingual tokens in closely related domains is crucial.  

As future work, we aim to develop a more training-robust and efficient pretraining paradigm to achieve domain knowledge acquisition.

\section{Limitations}
While this study provides novel insights into the mechanisms of bilingual domain adaptation, several limitations remain.
First, our analysis is based on a single case study involving English–Japanese biomedical adaptation. Although this setting offers a controlled environment to examine multilingual knowledge acquisition, the generalizability of the findings to other domains, language pairs, and model architectures remains to be validated. Future research should therefore extend the investigation to a broader range of large language models, including those trained under different configurations, as well as to diverse domain and linguistic contexts.

Second, our evaluation primarily relies on loss-based quantitative metrics to characterize the dynamics of knowledge acquisition and forgetting. While these measures provide a consistent and interpretable framework for tracking learning behavior, they do not fully capture how such knowledge is manifested in model outputs. A complementary qualitative and text-level evaluation, examining generated outputs and factual correctness, will be necessary to bridge the gap between internal training dynamics and externally observable performance.

\bibliography{acl_latex}

\appendix

\section{AdaXEval Implementation Details}

\subsection{Generation Details}
\label{appendix:adaxeval_setting}
In this section, we provide details on the process used to generate the AdaXEval dataset from J-STAGE documents. 

\smallskip\noindent\textbf{(1) Factual sentence filtering: } 
We use the open-source NLP tool HanLP\footnote{\url{https://github.com/hankcs/HanLP}}
 for Japanese sentence segmentation, and scispaCy\footnote{\url{https://github.com/allenai/scispacy}}~\cite{neumann-etal-2019-scispacy}
, which provides a full spaCy pipeline for scientific and biomedical texts, for English sentence segmentation. 
Subsequently, we perform biomedical named entity recognition (NER) on each sentence and filter out sentences containing fewer than two named entities. 
Specifically, for Japanese medical documents, we employ MedNERN-CR-JA\footnote{\url{https://github.com/sociocom/MedNERN-CR-JA}}~\cite{social_computing_lab_2023}
, a model specialized for NER in the Japanese medical domain. 
For English texts, we use the ``en\_ner\_bionlp13cg\_md'' model provided by scispaCy for biomedical entity extraction.

\smallskip\noindent\textbf{(2) Domain triple extraction: } 
In the next step, we use a multi-LLM agent to first judge whether the given sentence contains biomedical facts and extract them if it does. 
We carefully design CoT instruction with few-shot examples to instruct each LLM in the agent to generate a structural output, containing three fields: 
\begin{itemize}[leftmargin=1em,itemsep=1pt,topsep=2pt,parsep=1pt]
    \item \texttt{factuality}: answer \texttt{yes} or \texttt{no}. If \texttt{yes}, the model should output the \texttt{triple}; Otherwise, it outputs \texttt{None} to the \texttt{triple} field. 
    \item \texttt{triple}: A nested JSON object with three fields: subject, relation, and object representing the extracted fact. Please note that if factuality is false, make sure this field is null
    \item \texttt{reason}: A brief explanation for why the sentence was or wasn't considered factual, referring to the criteria provided. This is included to improve the model's reasoning ability.
\end{itemize}
The models' confidence in judging the factuality is measured by the probability of \texttt{yes} token out by the \texttt{factuality} field.

For our experiments, we utilize three strong open-source LLMs for both English and Japanese, opting for open-source models to avoid the high computational cost of commercial APIs. 
For \textbf{English biomedical triple extraction}, we employ Qwen-32B~\cite{qwen3technicalreport},\footnote{\url{https://huggingface.co/Qwen/Qwen3-32B}} DeepSeek-R1-Distill-Llama-70B~\cite{deepseekai2025deepseekr1incentivizingreasoningcapability},\footnote{\url{https://huggingface.co/deepseek-ai/DeepSeek-R1-Distill-Llama-70B}} and Llama-3.3-70B-Instruct~\cite{grattafiori2024llama3herdmodels}.\footnote{\url{https://huggingface.co/meta-llama/Llama-3.3-70B-Instruct}} 
For \textbf{Japanese biomedical triple extraction}, we use Qwen-32B, llm-jp-3.1-8x13b-instruct4,\footnote{\url{https://huggingface.co/llm-jp/llm-jp-3.1-8x13b-instruct4}} and Llama-3.3-Swallow-70B-Instruct-v0.4.\footnote{\url{https://huggingface.co/tokyotech-llm/Llama-3.3-Swallow-70B-Instruct-v0.4}} 
For each sentence, we aggregate the confidence scores from the three models and retain sentences with at least two models predicting a \texttt{yes} label. 
We then apply a heuristic method to select the final triple from the three candidates.

\smallskip\noindent\textbf{(3) Generation of queries and distractors: } 
We finally use the extracted triples and their corresponding context sentences as input to instruct the strong close-source LLM, GPT-4.1, for generating queries and distractors. 
We carefully design the prompts for both English and Japanese. 
The final generation is recorded and included in the final AdaXEval evaluation dataset. 

\smallskip\noindent\textbf{(4) Quality filtering:} 
Finally, we use LLM to conduct a two-steps quality filtering. 
First, we introduce a rigorous annotation instruction for assessing the quality of automatically generated biomedical knowledge questions.
Given a sentence, its knowledge triple, a fill-in-the-blank query, and a question-style paraphrase, we evaluate three aspects:
\begin{itemize}
[leftmargin=1em,itemsep=1pt,topsep=2pt,parsep=1pt]
    \item The fidelity and clarity of the cloze prompt,
    \item The semantic equivalence and self-containment of the paraphrased question, and
    \item The correctness and plausibility of the answer and distractors.
\end{itemize}
Then, we craft the annotation instruction to let LLM decide whether the document in the target languages contains the knowledge matching the created instances in the source languages. The annotation is to confirm the quality of the AdaXEval dataset in providing interlingual generalization evaluation.  

\subsection{Evaluation Metrics}
\label{appendix:adaxeval-evaluation}
We follow \cite{eval-harness} to compute the average cross-entropy loss over the target tokens of possible answers and select the one that has the highest generation possibility as the final answer. 
Specifically, for loss calculation of cloze queries, we use tokens before the \texttt{[BLANK]} as context and compute loss on the following tokens.
For paraphrases, we treat the question as context and measure only the loss of answer tokens.
We use prediction accuracy as the metric for knowledge acquisition. 
\smallskip\noindent\textbf{Formulation:}
Let the model be $p_\theta(\cdot \mid \cdot)$. 
Each dataset $\mathcal{D}$ contains pairs $(q, a)$, where $q$ is either a \emph{cloze prompt} or a \emph{paraphrase question}, and 
$a = (a_1, \dots, a_m)$ is the tokenized answer sequence (\textit{e.g.}, ``insulin''). 
For cloze prompts, the prompt contains a special token \texttt{[BLANK]}, and for paraphrases, the question is a natural question. 
We denote by $c$ the context tokens and by $s=(s_1,\dots,s_n)$ the evaluation sequence whose loss we measure. 

\smallskip\noindent\textbf{Cloze queries:}
For a cloze prompt such as:

\begin{raggedright}
\ttfamily
"[BLANK] can be used to control blood sugar level."
\end{raggedright}

\noindent{the evaluation sequence $s$ is the full completion after the \texttt{[BLANK]}, i.e.,}
\[
s = (a_1, \dots, a_m, r_1, \dots, r_k),
\]
where $a_1, \dots, a_m$ are answer tokens (``insulin''), and $r_1, \dots, r_k$ are the remainder tokens (`` can be used to control blood sugar level''). 
The average cross-entropy loss is defined as
\[
\mathcal{L}_{\text{cloze}}(q,a) = - \frac{1}{n} \sum_{t=1}^{n} \log p_\theta(s_t \mid c, s_{<t}).
\]

\smallskip\noindent\textbf{Paraphrase queries:}
For a paraphrase question such as

\begin{raggedright}
\ttfamily
"Which substance helps manage glycemic levels?"
\end{raggedright}

\noindent{we take the entire question tokens as context $c$, and the evaluation sequence is only the answer tokens:}
\[
s = (a_1, \dots, a_m).
\]
The loss is computed as
\[
\mathcal{L}_{\text{para}}(q,a) = - \frac{1}{m} \sum_{t=1}^{m} \log p_\theta(a_t \mid c, a_{<t}).
\]

\smallskip\noindent\textbf{Prediction and accuracy:}
For multiple-choice answers $\mathcal{A}=\{a^{(1)},\dots,a^{(K)}\}$, we select the candidate with the lowest loss (equivalently, highest likelihood):
\[
\hat{a} = \arg\min_{a \in \mathcal{A}} \mathcal{L}(q,a).
\]
Finally, the knowledge acquisition metric is accuracy:
\[
\text{Accuracy} = \frac{1}{|\mathcal{D}|} \sum_{(q,a) \in \mathcal{D}} \mathbf{1}[\hat{a}=a].
\]

\section{AdaXEval Dataset Details}
\label{appendix:adaxeval_examples}

\label{Statistics of AdaXEval Datasets}
\label{appendix:adaxeval-statistics}
We randomly sampled 10,000 parallel documents to generate the evaluation dataset. 
The number of instances after each step in AdaXEval generation is shown in Table~\ref{tab:adaxeval_stats}. 

\begin{table}[t]
    \small
    \centering
    \caption{Dataset statistics at each step of AdaXEval.}
    \label{tab:adaxeval_stats}
    \begin{tabular}{p{4cm}cc}
        \toprule
        \textbf{Step} & \textbf{English} & \textbf{Japanese}\\
        \midrule
        Sampled abstracts & 10,000 & 10,000\\
        Splitted sentences & 81,770 & 71,661\\
        Sentences after entity filtering ($\geq$2 entities) & 45,390 & 40,762 \\
        Triple extraction & 4840 & 3926 \\
        Cloze queries generation & 4840 & 3926 \\
        Paraphrases generation & 4840 & 3926 \\
        After Quality Filtering & 3236 & 2553 \\
        \bottomrule
    \end{tabular}
\end{table}

\subsection{Human Evaluation}
\label{appendix:human-evaluation}
To assess the quality of our generated datasets, we conduct a comprehensive human evaluation across four key components of the knowledge extraction and question generation pipeline. 
The annotation is conducted by the first author, who has a language background of both Japanese and English. For the annotation that requires specific domain knowledge, the author uses advanced LLMs, such as ChatGPT or Claude, as an assistant for annotation.
See the full human evaluation guideline documented in Appendix~\ref{appendix:human_eval_guide}.

\smallskip\noindent\textbf{(1) Cloze prompt evaluation} checks the faithfulness of the generated prompt to the original sentence structure. 

\smallskip\noindent\textbf{(2) Paraphrases evaluation} is conducted on four dimensions: fluency and grammaticality, linguistic diversity in reformulation, factual correctness to the original sentence in the source language, and interlingual factual correctness using the corresponding documents in the target language. 

\smallskip\noindent\textbf{(3) Distractor quality} is measured through plausibility within the domain and apparent incorrectness relative to the original context. 

Each metric employs structured scoring rubrics with scales ranging from 0-2 or 0-3, enabling systematic assessment of dataset quality across multiple linguistic and semantic dimensions.
We randomly sample 50 instances for each language and conduct human evaluation following the guidance above. 
As shown in Table~\ref{tab:human_eval_results}, our evaluation results indicate that AdaXEval is capable of generating high-quality evaluation data, meeting the requirements for assessing knowledge memorization as well as intralingual and interlingual generalization evaluation.

\begin{table}[t]
\small
\centering
\caption{Human evaluation results for knowledge acquisition datasets generated from the biomedical J-STAGE corpus.}
\begin{tabularx}{0.9\linewidth}{lcc}
\toprule
\textbf{Evaluation Metric} & \textbf{Japanese} & \textbf{English} \\
\hline
cloze prompt (Faithfulness) & 2.84/3 & 2.89/3 \\
\hline
Paraphrase (Fluency) & 2.94/3 & 2.96/3 \\
\hline
Paraphrase (Diversity) & 2.48/3 & 2.62/3 \\
\hline
Paraphrase (Factuality) & 2.86/3 & 2.86/3 \\
\hline
Paraphrase (Inter-Factuality) & 1.68/2 & 1.76/2 \\
\hline
Distractor (Plausibility) & 2.35/3 & 2.16/3 \\
\hline
Distractor (Incorrectness) & 2.89/3 & 2.97/3 \\
\hline
\end{tabularx}
\label{tab:human_eval_results}
\end{table}

\subsection{Examples}
\label{appendix:adaxeval-examples}
We randomly sample 10 examples from AdaXEval for both English and Japanese and display them in Table~\ref{tab:adaxeval-ja-samples} (Japanese) and Table~\ref{tab:adaxeval-en-samples} (English).

\section{Sequence Perturbation Analysis}
\label{appendix:sequence_perturb}

In this section, we introduce the detailed settings for sequence perturbation experiments and report the additional results. 

\subsection{Details of Perturbation}
For \emph{monosym@X} and \emph{monosym@X} that require collecting synonyms from WordNet 2.0, we only conduct Japanese-to-English replacement. 
Specifically, we first tokenize the Japanese sequence by sudachipy\footnote{\url{https://github.com/WorksApplications/SudachiPy}}, a Japanese morphological analyzer, and get the POS tags. 
Then we filter out tokens with stop words and the POS tags that are not \begin{CJK}{UTF8}{min} ``普通名詞'', ``固有名詞'', ``サ変接続'', ``形容動詞語幹'', ``動詞一般''\end{CJK} to avoid introducing noisy words. 
Furthermore, the paraphrasing and translation are done by requesting GPT-4.1. 

\subsection{Perturbation Results}
\begin{figure*}[htbp]
    \centering
    \subcaptionbox{\texttt{mask-X} with varied edit distance}[0.3\textwidth]{\includegraphics[width=0.3\textwidth]{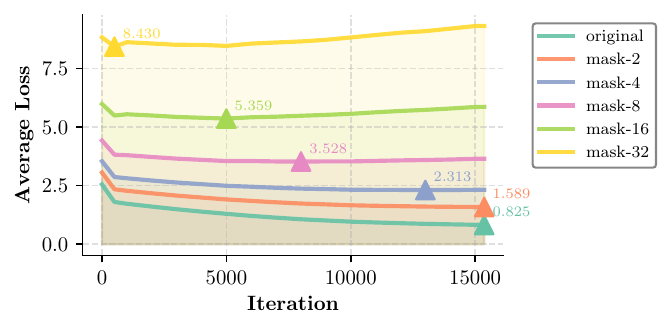}}
    \subcaptionbox{\texttt{random-X} with varied edit distance}[0.3\textwidth]{\includegraphics[width=0.3\textwidth]{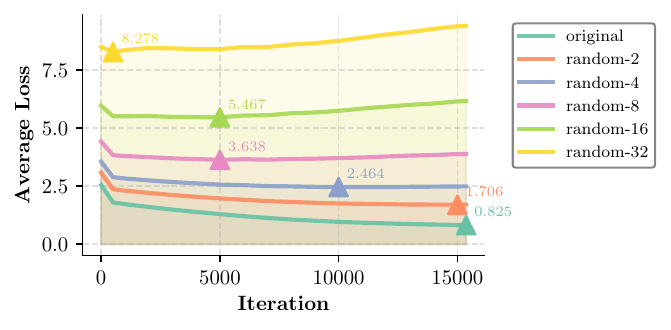}}
    \subcaptionbox{\texttt{reorder-X@3} with varied edit distance}[0.3\textwidth]{\includegraphics[width=0.3\textwidth]{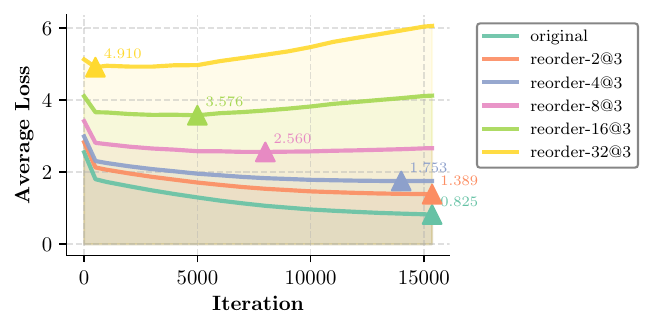}}
    \subcaptionbox{\texttt{delete-X} with varied edit distance}[0.3\textwidth]{\includegraphics[width=0.3\textwidth]{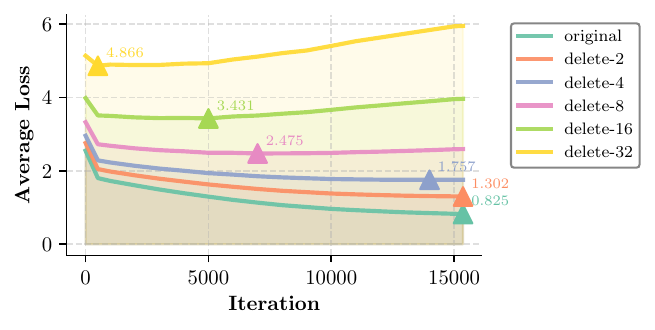}}
    \subcaptionbox{\texttt{monosyn-X} with varied edit distance}[0.3\textwidth]{\includegraphics[width=0.3\textwidth]{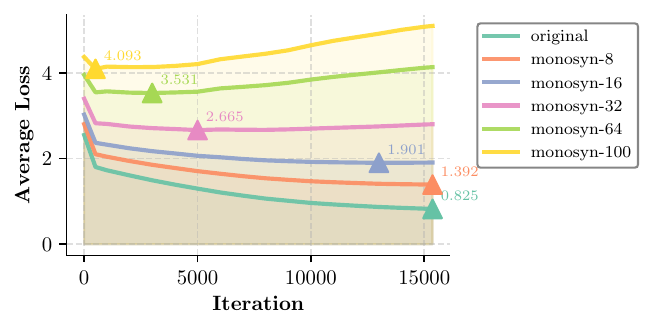}}
    \subcaptionbox{\texttt{mltlsyn-X} with varied edit distance}[0.3\textwidth]{\includegraphics[width=0.3\textwidth]{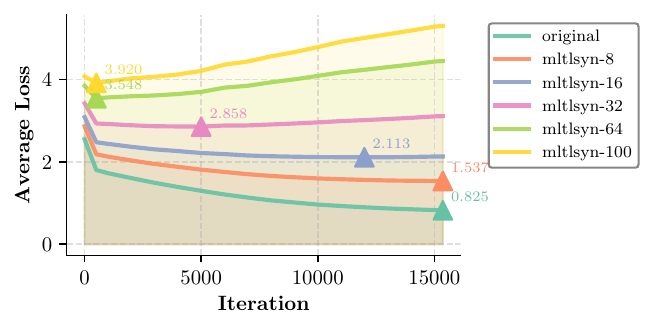}}
    \subcaptionbox{Token perturbation with 8\% distance}[0.3\textwidth]{\includegraphics[width=0.3\textwidth]{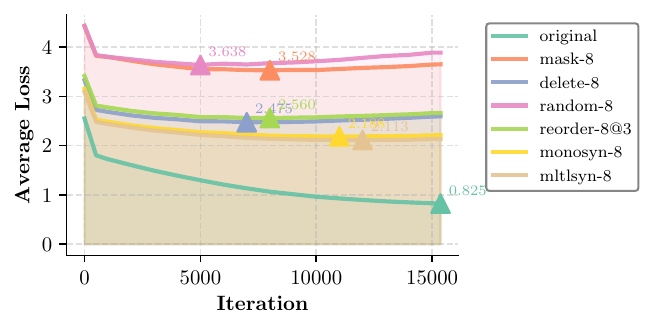}}
    \subcaptionbox{Token perturbation with 16\% distance}[0.3\textwidth]{\includegraphics[width=0.3\textwidth]{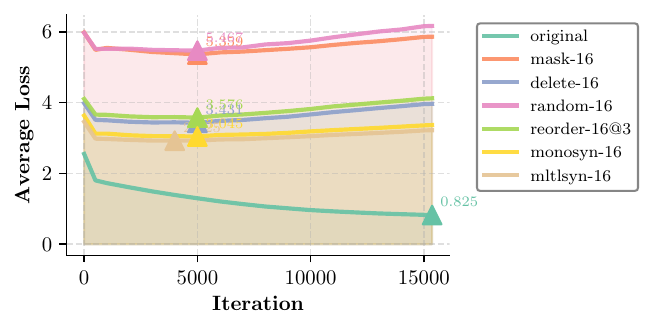}}
    \subcaptionbox{Token perturbation with 32\% distance}[0.3\textwidth]{\includegraphics[width=0.3\textwidth]{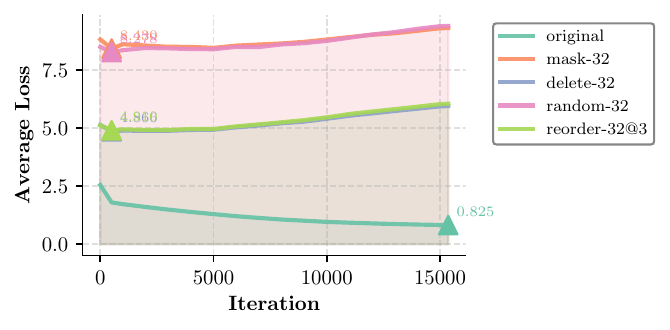}}
    
    \subcaptionbox{\texttt{partial-a}}[0.3\textwidth]{\includegraphics[width=0.3\textwidth]{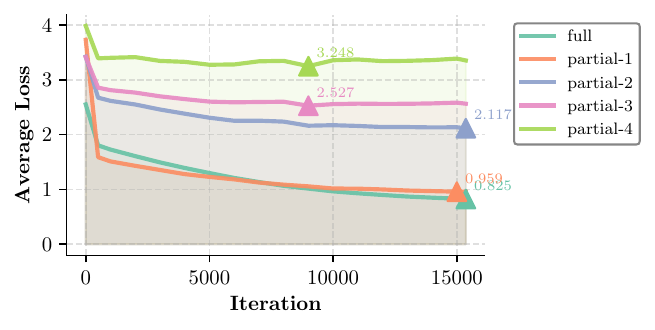}}
    \subcaptionbox{\texttt{partial-ab}}[0.3\textwidth]{\includegraphics[width=0.3\textwidth]{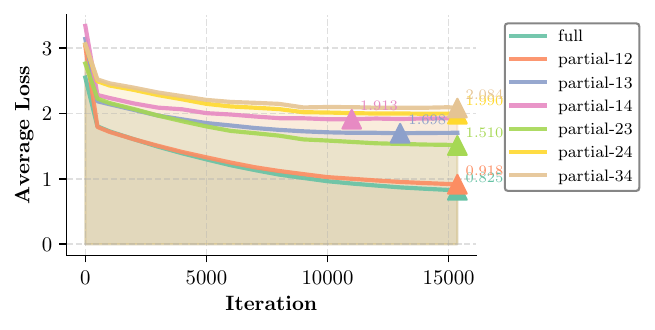}}
    \subcaptionbox{\texttt{syntax-X} with varied edit distance}[0.3\textwidth]{\includegraphics[width=0.3\textwidth]{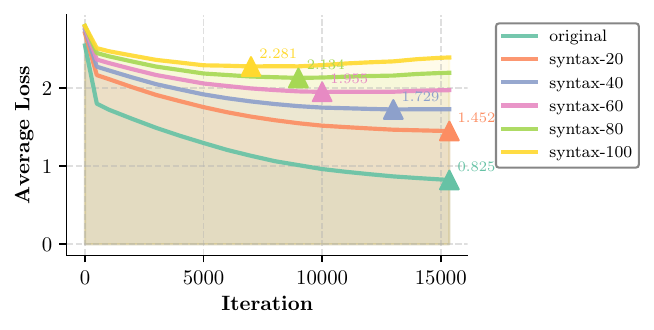}}
    \subcaptionbox{\texttt{lexicon-X} with varied edit distance}[0.3\textwidth]{\includegraphics[width=0.3\textwidth]{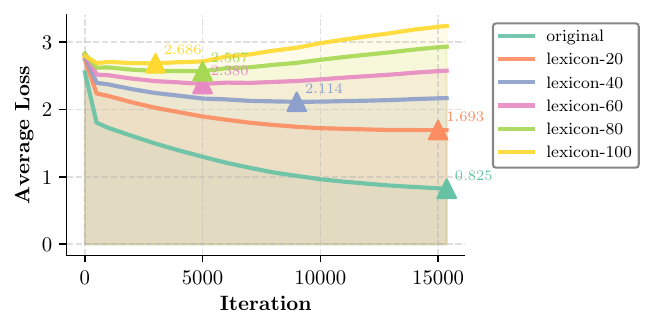}}
    \subcaptionbox{\texttt{semantic-X} with varied edit distance}[0.3\textwidth]{\includegraphics[width=0.3\textwidth]{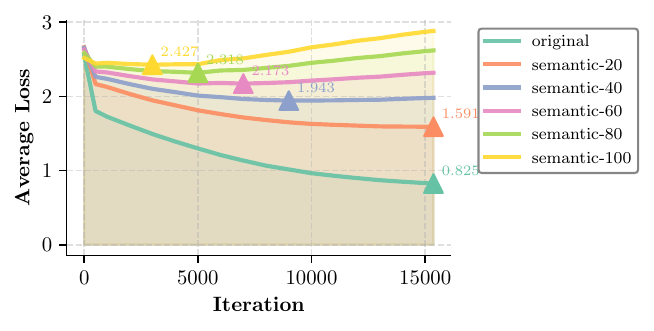}}
    \subcaptionbox{\texttt{trans-X} with varied edit distance}[0.3\textwidth]{\includegraphics[width=0.3\textwidth]{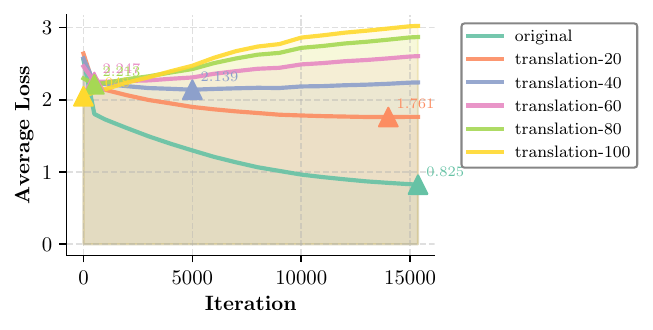}}
    \subcaptionbox{Sentence perturbation with 40\% edit distance}[0.3\textwidth]{\includegraphics[width=0.3\textwidth]{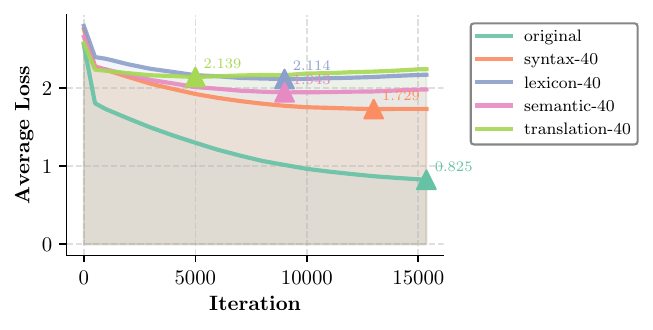}}
    \subcaptionbox{Sentence perturbation with 60\% edit distance}[0.3\textwidth]{\includegraphics[width=0.3\textwidth]{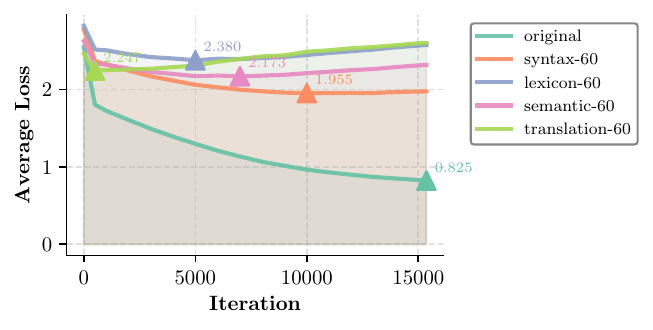}}
    \subcaptionbox{Sentence perturbation with 80\% edit distance}[0.3\textwidth]{\includegraphics[width=0.3\textwidth]{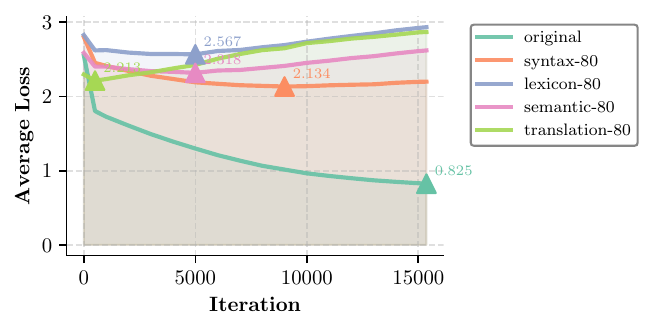}}
    \caption{The loss dynamics over all perturbation patterns on Japanese sequences.}
    \label{fig:ja_all_perturb}
\end{figure*}

\begin{figure*}[htbp]
    \centering
    \subcaptionbox{\texttt{mask-X} with varied edit distance}[0.3\textwidth]{\includegraphics[width=0.3\textwidth]{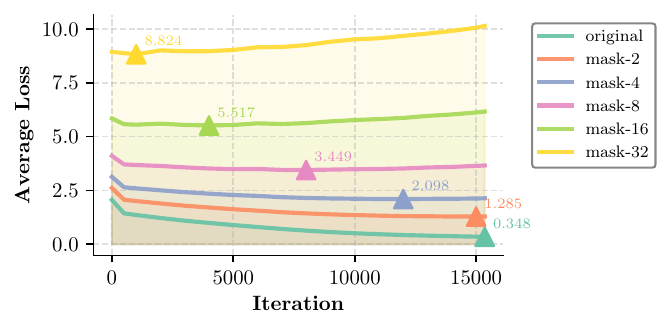}}
    \subcaptionbox{\texttt{random-X} with varied edit distance}[0.3\textwidth]{\includegraphics[width=0.3\textwidth]{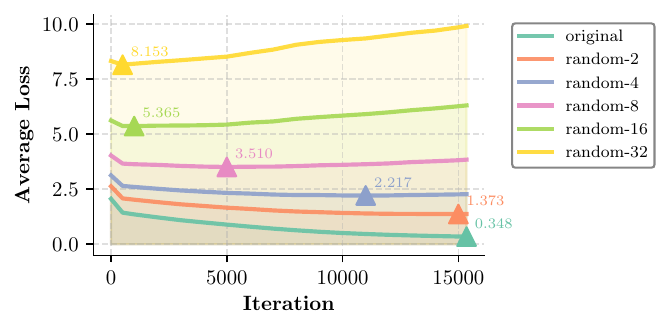}}
    \subcaptionbox{\texttt{reorder-X@3} with varied edit distance}[0.3\textwidth]{\includegraphics[width=0.3\textwidth]{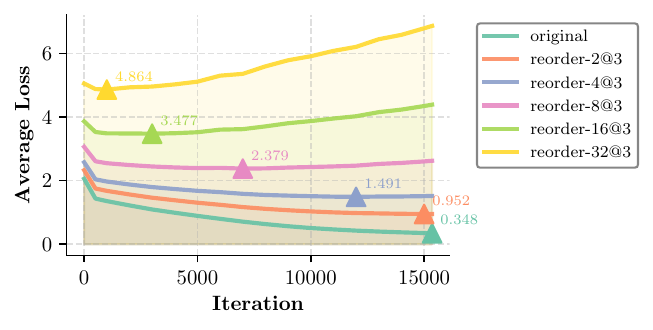}}
    \subcaptionbox{\texttt{delete-X} with varied edit distance}[0.3\textwidth]{\includegraphics[width=0.3\textwidth]{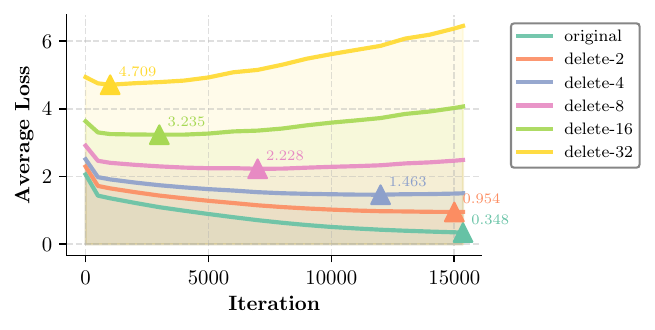}}
    \subcaptionbox{\texttt{monosyn-X} with varied edit distance}[0.3\textwidth]{\includegraphics[width=0.3\textwidth]{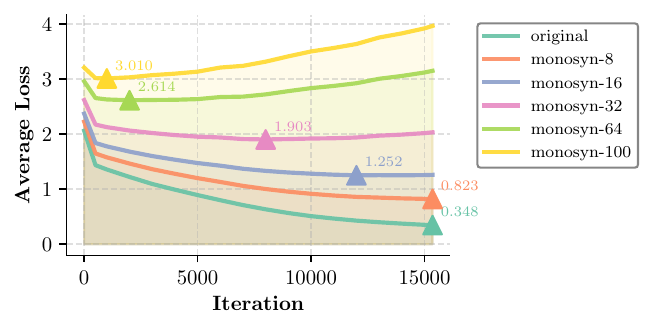}}
    \subcaptionbox{\texttt{mltlsyn-X} with varied edit distance}[0.3\textwidth]{\includegraphics[width=0.3\textwidth]{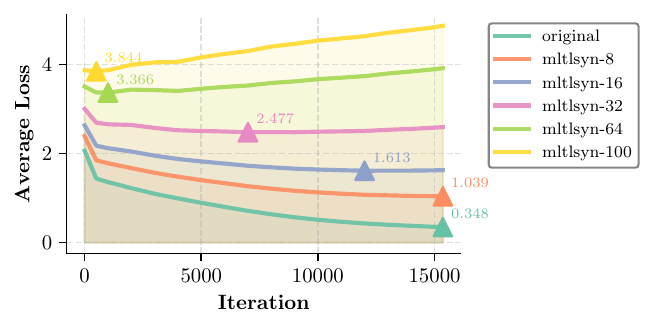}}
    \subcaptionbox{Token perturbation with 8\% distance}[0.3\textwidth]{\includegraphics[width=0.3\textwidth]{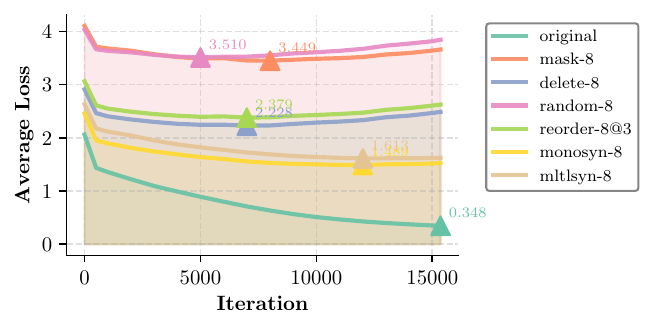}}
    \subcaptionbox{Token perturbation with 16\% distance}[0.3\textwidth]{\includegraphics[width=0.3\textwidth]{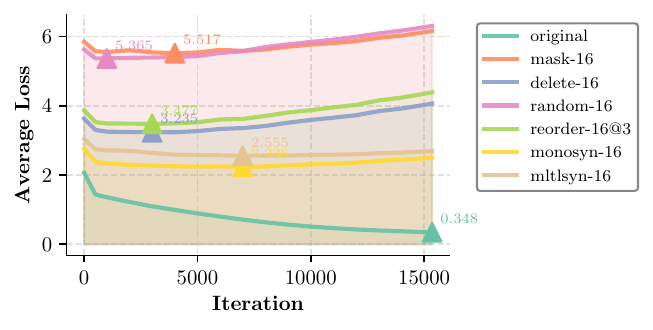}}
    \subcaptionbox{Token perturbation with 32\% distance}[0.3\textwidth]{\includegraphics[width=0.3\textwidth]{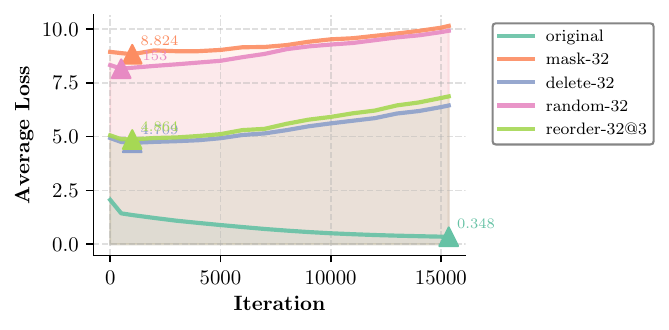}}
    
    \subcaptionbox{\texttt{partial-a}}[0.3\textwidth]{\includegraphics[width=0.3\textwidth]{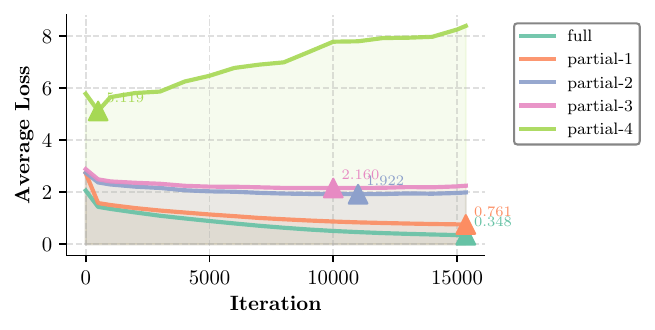}}
    \subcaptionbox{\texttt{partial-ab}}[0.3\textwidth]{\includegraphics[width=0.3\textwidth]{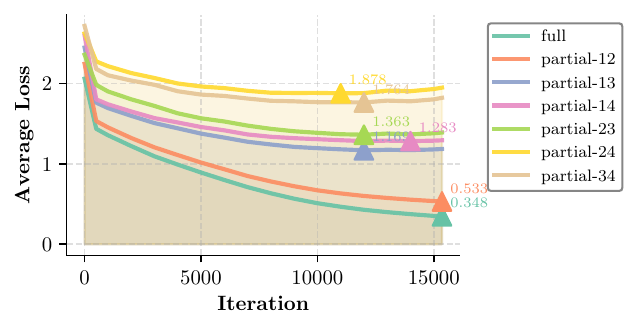}}
    \subcaptionbox{\texttt{syntax-X} with varied edit distance}[0.3\textwidth]{\includegraphics[width=0.3\textwidth]{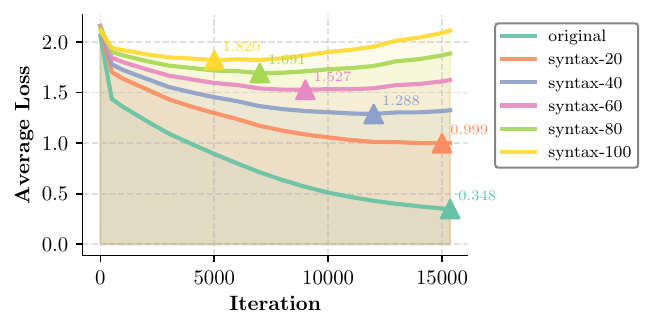}}
    \subcaptionbox{\texttt{lexicon-X} with varied edit distance}[0.3\textwidth]{\includegraphics[width=0.3\textwidth]{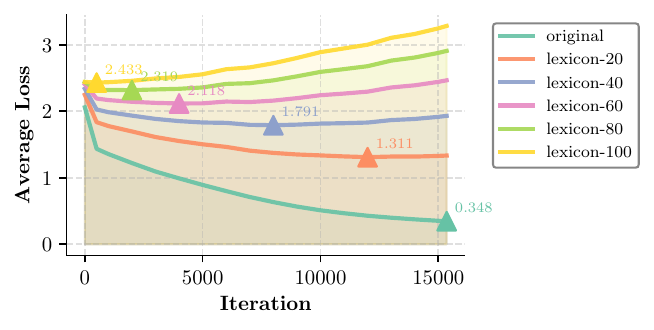}}
    \subcaptionbox{\texttt{semantic-X} with varied edit distance}[0.3\textwidth]{\includegraphics[width=0.3\textwidth]{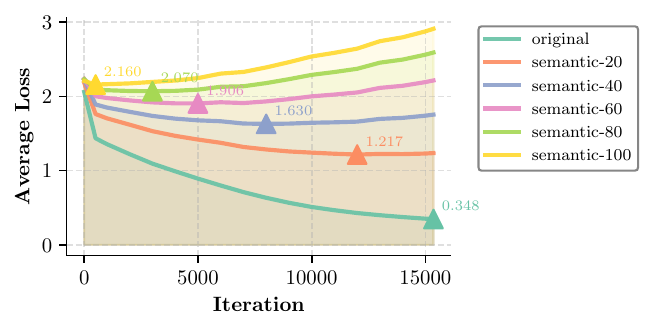}}
    \subcaptionbox{\texttt{trans-X} with varied edit distance}[0.3\textwidth]{\includegraphics[width=0.3\textwidth]{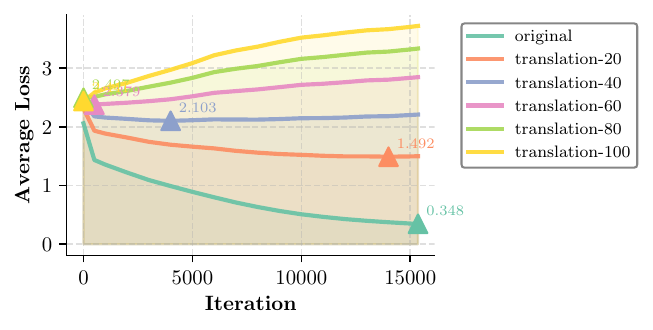}}
    \subcaptionbox{Sentence perturbation with 40\% edit distance}[0.3\textwidth]{\includegraphics[width=0.3\textwidth]{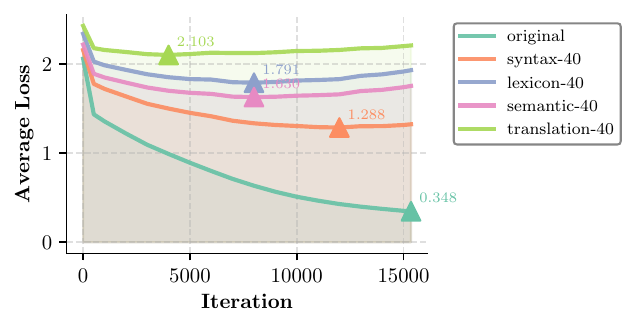}}
    \subcaptionbox{Sentence perturbation with 60\% edit distance}[0.3\textwidth]{\includegraphics[width=0.3\textwidth]{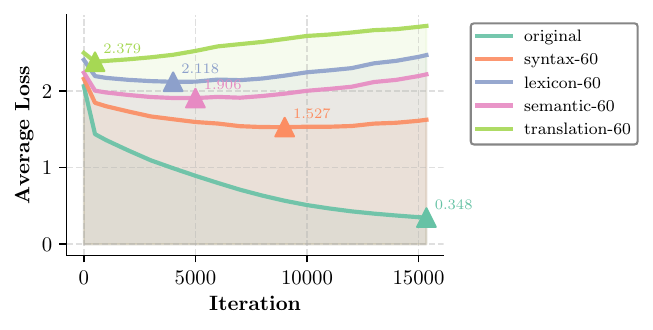}}
    \subcaptionbox{Sentence perturbation with 80\% edit distance}[0.3\textwidth]{\includegraphics[width=0.3\textwidth]{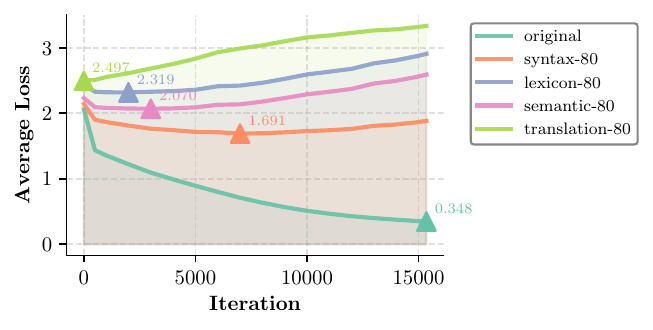}}
    \caption{The loss dynamics over all perturbation patterns on English sequences.}
    \label{fig:en_all_perturb}
\end{figure*}

We show all the token-level perturbation results in Figure~\ref{fig:ja_all_perturb} and sentence-level perturbation result in Figure~\ref{fig:en_all_perturb}. 



\section{Domain Adaptation Training Details}
\label{appendix:training_settings}

\subsection{Training Data Generation}
\label{appendix:data_generation}
In this study, to address the scarcity of bilingual domain corpora and enhance domain understanding, we employ two data augmentation strategies: regex-based pattern mining and LLM-based QA generation. Both approaches yield instruction-like sequences, which we mix with raw corpora. Following prior work~\cite{cheng-etal-2024-instruction}, we adopt an instruction-pretraining strategy for continual domain adaptation.

\smallskip\noindent\textbf{(1) Regex-based pattern mining: }
\cite{instr-pretrain-qa} verified that by transforming raw corpora into reading comprehension texts, continual training can consistently enhance performance across various tasks in different domains. We adopt a similar strategy by analyzing the training corpora and mining regex patterns to automatically create instruction-style data. 
Furthermore, to increase data diversity, we prepare ten instruction templates for each type of reading comprehension text. For each document, however, we sample only one template per type.

Specifically, each document in J-STAGE contains multiple metadata fields, including: 
\begin{itemize}
    \item \texttt{title}: the title of the paper 
    \item \texttt{abstract}: the paper abstract
    \item \texttt{keywords}: pre-defined keywords of the paper
    \item \texttt{fields}: research categories of the paper
\end{itemize}

Based on this information, we construct ten types of reading comprehension instructions as follows: 
\begin{itemize}[leftmargin=1em,itemsep=0pt,topsep=1pt,parsep=0pt]
    \item \textbf{Summarization}: Summarize the context into one concise sentence, taking the abstract as input and the title as output.
    \item \textbf{Keyword Extraction}: Extract the keywords from the abstract, using the \texttt{keywords} field as the gold reference. 
    \item \textbf{Field Identification}: Identify the research field(s) of the paper, taking the abstract as input and the \texttt{fields} metadata as the expected output. 
    \item \textbf{Translation}: Translate between English and Japanese, using bilingual metadata or parallel text segments as input-output pairs. 
    \item \textbf{Text Completion}: Complete an incomplete abstract or title given the partial text, where the remainder of the text serves as the reference output. 
    \item \textbf{Conclusion Derivation}: Derive the study’s conclusion from its context, with the conclusion section as supervision. 
    \item \textbf{Background Derivation}: Infer the background or motivation of the study from the provided abstract or introduction sentences. 
    \item \textbf{Diagnosis}: Given a description of symptoms (extracted from biomedical corpora), predict the corresponding diagnosis, using annotated datasets where available. 
    \item \textbf{Reordering}: Reorder shuffled sentences into their natural sequence, ensuring coherence with the original abstract or section structure. 
    \item \textbf{Goal–Method–Result–Conclusion (GMRC)}: Derive one missing component (\textit{e.g.}, goal, method, result, or conclusion) based on the other three, enabling comprehension of scientific discourse structures. 
\end{itemize}

\smallskip\noindent\textbf{(2) LLM-base QA generation: }
To enhance the data diversity, we further generate five question-answer pairs for each document.
Specifically, we use DeepSeek-R1-Distill-Llama-70B for English QA-pair generation and use DeepSeek-R1-Distill-Qwen-JP-32B~\cite{cyberagent-deepseek-r1-distill-qwen-32b-japanese}\footnote{\url{https://huggingface.co/cyberagent/DeepSeek-R1-Distill-Qwen-32B-Japanese}} for Japanese QA generation. 
Noted that not all documents can successfully generate five QA pairs.

\subsection{continual training Settings}
\label{appendix:training_parameter}

We conduct continual training using \textbf{Megatron-LM} on the \texttt{llm-jp-3-13B} model. 
The training setup follows a distributed configuration with 4 compute nodes, each equipped with 8 A100 GPUs. 
We apply a tensor parallel size of 2 and a pipeline parallel size of 4, enabling efficient large-scale training with a sequence length of 4096. 
The optimizer is configured with a learning rate of $2\times10^{-5}$, weight decay of 0.1, and gradient clipping of 1.0, with a minimum learning rate of $2\times10^{-6}$. 
We adopt a micro-batch size of 1 and a global batch size of 32 to stabilize training. 
Under this configuration, training one epoch on 0.5B tokens requires approximately 7 hours, demonstrating the computational feasibility of continual training while maintaining efficiency on large-scale biomedical and cross-lingual corpora.

\section{Transfer From Japanese to English}
\label{appendix:cross-lingual-transfer-evaluation}

\begin{figure}[t]
  \centering
  \includegraphics[width=\linewidth]{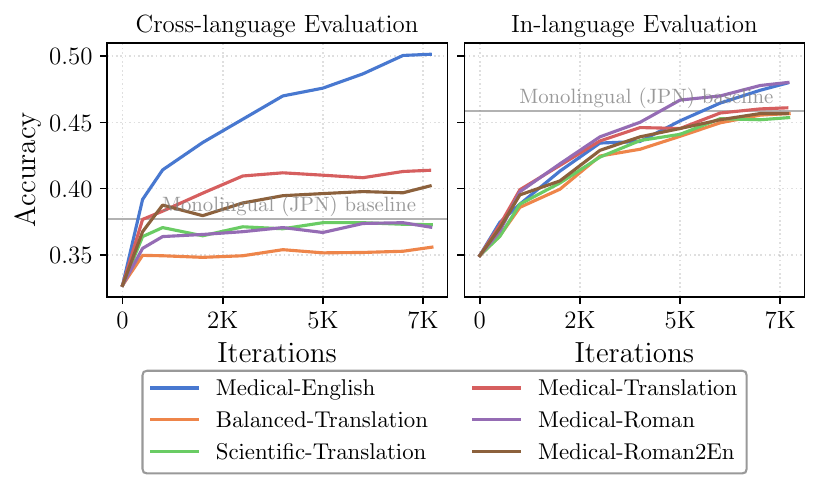}
  \caption{Japanese-to-English transfer evaluation with diverse strategies.}
  \label{fig:ja2en-train-strategies}
\end{figure}

Here, we report the results of the Japanese-to-English cross-lingual transfer evaluation using different recipes, as a supplement to the analysis in \S~\ref{subsec:cross-lingual-transfer-evaluation}. 
Specifically, in this setting, $\mathcal{C_K}$ is composed of the Japanese monolingual training corpora, while we vary the data source of $\mathcal{C_T}$ to investigate efficient cross-lingual transfer.
Evaluation results are shown in Figure~\ref{fig:ja2en-train-strategies}.

The evaluation results shown in Figure~\ref{fig:ja2en-train-strategies} exhibit trends that differ from those in Figure~\ref{fig:crosslingual-transfer-enhancement}, as the strong baseline using multilingual corpora (a mixture of Japanese and English monolingual training corpora) largely outperforms other translation corpora. 
This suggests that generalization within English knowledge is an easier task than transferring knowledge from Japanese to English. 
Nevertheless, we still observe the strongest cross-lingual transfer in the closely related domain compared to the other three domains.

\section{AdaXEval Human Evaluation Guideline}
\label{appendix:human_eval_guide}

\subsection{Cloze prompt Quality Evaluation}

\smallskip\noindent\textbf{Objective:}

Evaluate whether fill-in-the-blank prompts generated from biomedical academic sentences and subject–relation–object triples are clear, faithful to the original sentence, and free from factual distortion.
Each prompt removes the object from the triple and replaces it with a blank.

\smallskip\noindent\textbf{Input:}
You will be shown the following for each item:

\begin{itemize}[leftmargin=1em,itemsep=0pt,topsep=1pt,parsep=0pt]
    \item Original Sentence: The full academic sentence from which the triple is extracted.
    \item Triple: A \texttt{<subject, relation, object>} triple derived from that sentence.
    \item Generated Prompt: A fill-in-the-blank sentence where the object is replaced with \texttt{[BLANK]}.
\end{itemize}

\smallskip\noindent\textbf{Faithfulness Criteria}

When the \texttt{[BLANK]} is replaced with the object, does the prompt preserve the structure and meaning of the original sentence, including key contextual information?

\begin{itemize}[leftmargin=1em,itemsep=0pt,topsep=1pt,parsep=0pt]
    \item \textbf{What it checks:} Structural and contextual similarity between the prompt and the original sentence
    \item \textbf{Focuses on:} Wording, phrasing, sentence structure, presence of supporting context
\end{itemize}

\begin{table}[t]
\small
\centering
\begin{tabularx}{0.9\linewidth}{cX}
\toprule
\textbf{Score} & \textbf{Description} \\
\midrule
3 & The prompt closely mirrors the original sentence's structure and meaning. Minor surface-level changes (\textit{e.g.}, auxiliary verbs, punctuation, or sentence breaks) are acceptable. \\
\hline
2 & The core meaning is preserved, but there are moderate changes in wording, omission that does not change the semantic meaning, or noticeable rephrasing. \\
\hline
1 & The prompt differs significantly in structure or phrasing, or some key information is missed \\
\hline
0 & The prompt is not clearly based on the original sentence or appears unrelated in meaning or form. \\
\hline
\end{tabularx}
\end{table}

\subsection{Paraphrased Question Quality Evaluation}

\smallskip\noindent\textbf{Objective:}
Evaluate whether \textbf{question-style prompts}, automatically generated from fill-in-the-blank biomedical prompts, are:

\begin{itemize}[leftmargin=1em,itemsep=0pt,topsep=1pt,parsep=0pt]
    \item Semantically faithful to the original prompt (i.e., same question being asked)
    \item Grammatically correct and fluent
    \item Natural as questions a human would realistically ask
\end{itemize}

\smallskip\noindent\textbf{Input:}

\begin{itemize}[leftmargin=1em,itemsep=0pt,topsep=1pt,parsep=0pt]
    \item Original Fill-in-the-Blank Prompt (\textit{e.g.}, ``EGFR is highly expressed in [BLANK].'')
    \item Paraphrased Question-style Prompt (\textit{e.g.}, ``In which condition is EGFR highly expressed?'')
\end{itemize}

\smallskip\noindent\textbf{Example:}

\begin{itemize}[leftmargin=1em,itemsep=0pt,topsep=1pt,parsep=0pt]
    \item Original Sentence: ``EGFR is highly expressed in non-small cell lung carcinoma.''
    \item Triple: \textit{(EGFR, is highly expressed in, non-small cell lung carcinoma)}
    \item Prompt: ``EGFR is highly expressed in [BLANK].''
\end{itemize}

\smallskip\noindent\textbf{Evaluation Criteria}

\smallskip\noindent\textbf{1) Fluency and Grammaticality:}
Is the question grammatically correct, fluent, and natural-sounding in English?

\begin{itemize}[leftmargin=1em,itemsep=0pt,topsep=1pt,parsep=0pt]
    \item Focuses on syntax, awkward phrasing, unnatural interrogative forms
\end{itemize}

\begin{table}[t]
\small
\centering
\begin{tabularx}{0.9\linewidth}{cX}
\toprule
\textbf{Score} & \textbf{Description} \\
\midrule
3 & Fully natural and fluent; well-formed question \\
\hline
2 & Mostly fluent; minor grammatical issues or slight awkwardness \\
\hline
1 & Understandable but ungrammatical or clearly unnatural \\
\hline
0 & Ungrammatical, confusing, or not a valid question \\
\hline
\end{tabularx}
\end{table}

\smallskip\noindent\textbf{2) Factual Consistency:}
When given the answer, is the question factually consistent with the original sentence and triple (no distortion of meaning or relationships)?

\begin{itemize}[leftmargin=1em,itemsep=0pt,topsep=1pt,parsep=0pt]
    \item \textbf{Semantic Accuracy:} Does the question preserve the intended meaning of the original sentence and triple?
    \item \textbf{Relationship Integrity:} Are the logical relations (e.g., cause/effect, association, identity) between subject, relation, and object kept intact?
    \item \textbf{Context Preservation:} Does the question retain essential context needed to uniquely identify the correct answer (e.g., disease type, anatomical site, conditions, time point, numerical thresholds)? Missing such context should lower the score.
\end{itemize}

\begin{table}[t]
\small
\centering
\begin{tabularx}{0.9\linewidth}{cX}
\toprule
\textbf{Score} & \textbf{Description} \\
\midrule
3 & The question fully matches the meaning of the original facts; no distortion or loss of critical details; context is complete and the answer remains uniquely correct. \\
\hline
2 & The question is generally accurate but has minor factual ambiguity or slightly softer interpretation that could cause mild uncertainty while still pointing to the same answer. \\
\hline
1 & Some key context or factual precision is missing or altered; relationships are weakened; the answer is still inferable but not strictly unique or reliable. \\
\hline
0 & The question introduces clear factual errors, distorts the original meaning or relationships, or misleads so the answer could be wrong or invalid. \\
\hline
\end{tabularx}
\end{table}

\smallskip\noindent\textbf{3) Linguistic Diversity:}
How well does the paraphrased question use different wording and structure from the original prompt?

\begin{itemize}[leftmargin=1em,itemsep=0pt,topsep=1pt,parsep=0pt]
    \item \textbf{What it checks:} Lexical and syntactic variation between the original and paraphrased versions
    \item \textbf{Focuses on:} Synonym usage, sentence structure changes, reformulation techniques
\end{itemize}

\begin{table}[t]
\small
\centering
\begin{tabularx}{0.9\linewidth}{cX}
\toprule
\textbf{Score} & \textbf{Description} \\
\midrule
3 & Excellent reformulation; uses different vocabulary and structure while maintaining meaning \\
\hline
2 & Good variation; some different wording but follows similar structure or only changing the structure without introducing new vocabulary \\
\hline
1 & Minimal variation; mostly replaces the blank with a question word \\
\hline
0 & No meaningful reformulation; essentially the same as the original with a question mark \\
\hline
\end{tabularx}
\end{table}

\smallskip\noindent\textbf{4) Cross-Lang Factual Consistency:}
Can the evidence supporting the same fact be found or inferred from the document in the target language?

\begin{table}[t]
\small
\centering
\begin{tabularx}{0.9\linewidth}{cX}
\toprule
\textbf{Score} & \textbf{Description} \\
\midrule
2 & Direct match -- same triple is clearly expressed in one sentence or consecutive sentences in the abstract in the target language. \\
\hline
1 & Inferable without extra biomedical knowledge out of the given content -- not in one sentence, but can be reasonably inferred from the paragraph as a whole. \\
\hline
0 & Not supported or requires extra knowledge -- the triple cannot be inferred from the abstract, or unrelated. \\
\hline
\end{tabularx}
\end{table}

\subsection{Distractor Quality Evaluation}

\smallskip\noindent\textbf{Objective:}
Evaluate the quality of three distractor options (incorrect candidates) accompanying the correct answer (object) in a multiple-choice setting derived from biomedical fill-in-the-blank prompts.
Each distractor should be:

\begin{itemize}[leftmargin=1em,itemsep=0pt,topsep=1pt,parsep=0pt]
    \item Plausible given the question
    \item Incorrect (not the original object)
    \item Relevant in context and domain
\end{itemize}

\smallskip\noindent\textbf{Input:}

\begin{itemize}[leftmargin=1em,itemsep=0pt,topsep=1pt,parsep=0pt]
    \item Original Sentence
    \item Triple
    \item Fill-in-the-Blank Prompt
    \item Answer options
\end{itemize}

\smallskip\noindent\textbf{Example:}

\begin{itemize}[leftmargin=1em,itemsep=0pt,topsep=1pt,parsep=0pt]
    \item Original Sentence: \texttt{EGFR is highly expressed in non-small cell lung carcinoma.}
    \item Triple: \textit{(EGFR, is highly expressed in, non-small cell lung carcinoma)}
    \item Prompt: \texttt{EGFR is highly expressed in [BLANK].}
\end{itemize}

\smallskip\noindent\textbf{Evaluation Criteria:}
Need evaluations for both the cloze prompt and the paraphrased question.

\smallskip\noindent\textbf{1) Plausibility in Context:}
Is the distractor believable given the prompt and domain knowledge (biomedical)?

\begin{itemize}[leftmargin=1em,itemsep=0pt,topsep=1pt,parsep=0pt]
    \item \textbf{What it checks:} subject, relation, and expected answer type (should be the object)
    \item \textbf{Focuses on:} Be cautious of meaning shifts, incorrect substitutions, or role reversals.
\end{itemize}

\begin{table}[t]
\small
\centering
\begin{tabularx}{0.9\linewidth}{cX}
\toprule
\textbf{Score} & \textbf{Description} \\
\midrule
3 & Highly plausible: Very convincing as an answer; can confuse even experts; fits subject, relation, domain well. \\
\hline
2 & Moderately plausible: Makes sense in general; fits domain and context somewhat; can be ruled out by basic domain knowledge. \\
\hline
1 & Barely plausible: Awkward or uncommon; easily ruled out by surface cues or common sense without any domain knowledge. \\
\hline
0 & Implausible: Irrelevant, nonsensical, or grammatically incorrect; not a valid answer option. \\
\hline
\end{tabularx}
\end{table}

\smallskip\noindent\textbf{2) Incorrectness:}
Is the distractor clearly incorrect given the original sentence and triple?

\begin{table}[t]
\small
\centering
\begin{tabularx}{0.9\linewidth}{cX}
\toprule
\textbf{Score} & \textbf{Description} \\
\midrule
3 & Definitely wrong: contradicts or is not supported by the original sentence. \\
\hline
2 & Likely wrong: but could be ambiguous or partially true given the original sentence. \\
\hline
1 & Borderline: Possibly true or partially correct; ambiguous given the sentence. \\
\hline
0 & Incorrectly labeled -- This distractor is actually correct or the original answer given the original sentence. \\
\hline
\end{tabularx}
\end{table}

\begin{table*}[htbp]
\scriptsize
\centering
\renewcommand{\arraystretch}{1.3} 
\caption{Samples of Japanese AdaXEval Dataset.}
\begin{tabularx}{\textwidth}{X X X p{5cm} c}
\hline
\textbf{Sentence} & \textbf{cloze prompt} & \textbf{Paraphrase} & \textbf{Options} & \textbf{Answer ID} \\
\hline
\begin{CJK}{UTF8}{min} その結果, Ca-HApのカルシウムおよびリンの溶解は多くの有機酸水溶液中, pH4~6.5の範囲で下記の式に従うことがわかった (ただし, リンゴ酸, 酒石酸, クエン酸を除く) 。 \end{CJK} & \begin{CJK}{UTF8}{min} その結果、[BLANK]は多くの有機酸水溶液中、pH4〜6.5の範囲で下記の式に従うことがわかった（ただし、リンゴ酸、酒石酸、クエン酸を除く）。 \end{CJK} & \begin{CJK}{UTF8}{min} 多くの有機酸水溶液中、pH4〜6.5の範囲で下記の式に従うことがわかったのは何の溶解ですか？ \end{CJK} & \begin{CJK}{UTF8}{min} A. Ca-HApのマグネシウムおよび鉄の溶解 \end{CJK} \newline
\begin{CJK}{UTF8}{min} B. Ca-HApの亜鉛および銅の溶解 \end{CJK} \newline
\begin{CJK}{UTF8}{min} C. Ca-HApのカルシウムおよびリンの溶解 \end{CJK} \newline
\begin{CJK}{UTF8}{min} D. Ca-HApのナトリウムおよびカリウムの溶解 \end{CJK} & \begin{CJK}{UTF8}{min} C \end{CJK} \\\hline
\begin{CJK}{UTF8}{min} それとともに, β-lactamaseやaminoglycoside acetyltransferase (AAC)などによる抗菌薬の不活化やDNA gyraseの変化による抗菌薬親和性の減少, さらにbiofilm形成による抗菌薬の低浸透などが組み合わさり多剤耐性化する. \end{CJK} & \begin{CJK}{UTF8}{min} β-lactamaseやaminoglycoside acetyltransferase (AAC)による抗菌薬の不活化、DNA gyraseの変化による抗菌薬親和性の減少、さらにbiofilm形成による抗菌薬の低浸透などが組み合わさり[BLANK]する。 \end{CJK} & \begin{CJK}{UTF8}{min} β-lactamaseやAACによる抗菌薬の不活化、DNA gyraseの変化、biofilm形成などが組み合わさることで生じる現象は何ですか？ \end{CJK} & \begin{CJK}{UTF8}{min} A. 単剤耐性化 \end{CJK} \newline
\begin{CJK}{UTF8}{min} B. 多剤耐性化 \end{CJK} \newline
\begin{CJK}{UTF8}{min} C. 抗菌薬感受性の増加 \end{CJK} \newline
\begin{CJK}{UTF8}{min} D. 抗菌薬の副作用増強 \end{CJK} & \begin{CJK}{UTF8}{min} B \end{CJK} \\\hline
\begin{CJK}{UTF8}{min} 悪性腫瘍に伴う血液凝固能亢進状態により脳卒中をきたす病態はTrousseau症候群として知られている. \end{CJK} & \begin{CJK}{UTF8}{min} 悪性腫瘍に伴う血液凝固能亢進状態により脳卒中をきたす病態は[BLANK]として知られている。 \end{CJK} & \begin{CJK}{UTF8}{min} 悪性腫瘍に伴う血液凝固能亢進状態が原因で脳卒中を発症する病態は何と呼ばれていますか？ \end{CJK} & \begin{CJK}{UTF8}{min} A. Goodpasture症候群 \end{CJK} \newline
\begin{CJK}{UTF8}{min} B. Trousseau症候群 \end{CJK} \newline
\begin{CJK}{UTF8}{min} C. 抗リン脂質抗体症候群 \end{CJK} \newline
\begin{CJK}{UTF8}{min} D. Lambert-Eaton症候群 \end{CJK} & \begin{CJK}{UTF8}{min} B \end{CJK} \\\hline
\begin{CJK}{UTF8}{min} 根部よりの吸収は化合物の疎水性 (logP) と負の相関性を示した. \end{CJK} & \begin{CJK}{UTF8}{min} [BLANK]は化合物の疎水性 (logP) と負の相関性を示した。 \end{CJK} & \begin{CJK}{UTF8}{min} 化合物の疎水性 (logP) と負の相関性を示すのはどのような吸収ですか？ \end{CJK} & \begin{CJK}{UTF8}{min} A. 根毛からの吸収 \end{CJK} \newline
\begin{CJK}{UTF8}{min} B. 茎部よりの吸収 \end{CJK} \newline
\begin{CJK}{UTF8}{min} C. 葉部よりの吸収 \end{CJK} \newline
\begin{CJK}{UTF8}{min} D. 根部よりの吸収 \end{CJK} & \begin{CJK}{UTF8}{min} D \end{CJK} \\\hline
\begin{CJK}{UTF8}{min} このことからp53はH2AXリン酸化には関与しておらず、脱リン酸化やさらに下流の因子と関わってあると考えられる。 \end{CJK} & \begin{CJK}{UTF8}{min} このことから[BLANK]はH2AXリン酸化には関与しておらず、脱リン酸化やさらに下流の因子と関わってあると考えられる。 \end{CJK} & \begin{CJK}{UTF8}{min} H2AXリン酸化に関与していないと考えられるタンパク質は何ですか？ \end{CJK} & \begin{CJK}{UTF8}{min} A. DNA-PK \end{CJK} \newline
\begin{CJK}{UTF8}{min} B. CHK2 \end{CJK} \newline
\begin{CJK}{UTF8}{min} C. p53 \end{CJK} \newline
\begin{CJK}{UTF8}{min} D. ATM \end{CJK} & \begin{CJK}{UTF8}{min} C \end{CJK} \\\hline
\begin{CJK}{UTF8}{min} Solitary fibrous tumor(SFT)は,間葉系細胞由来の稀な腫瘍である. \end{CJK} & \begin{CJK}{UTF8}{min} [BLANK]は、間葉系細胞由来の稀な腫瘍である。 \end{CJK} & \begin{CJK}{UTF8}{min} 間葉系細胞由来の稀な腫瘍として知られているのは何ですか？ \end{CJK} & \begin{CJK}{UTF8}{min} A. 神経膠腫 (glioma) \end{CJK} \newline
\begin{CJK}{UTF8}{min} B. リンパ腫 (lymphoma) \end{CJK} \newline
\begin{CJK}{UTF8}{min} C. Solitary fibrous tumor (SFT) \end{CJK} \newline
\begin{CJK}{UTF8}{min} D. 扁平上皮癌 (squamous cell carcinoma) \end{CJK} & \begin{CJK}{UTF8}{min} C \end{CJK} \\\hline
\begin{CJK}{UTF8}{min} 1990年代半ばに開発された脱窒菌法により,硝酸イオンのδ15N,δ18Oを微量で同時に測定できるようになった。 \end{CJK} & \begin{CJK}{UTF8}{min} 1990年代半ばに開発された[BLANK]により、硝酸イオンのδ15N,δ18Oを微量で同時に測定できるようになった。 \end{CJK} & \begin{CJK}{UTF8}{min} 硝酸イオンのδ15Nおよびδ18Oを微量で同時に測定することを可能にした方法は何ですか？ \end{CJK} & \begin{CJK}{UTF8}{min} A. 分光光度法 \end{CJK} \newline
\begin{CJK}{UTF8}{min} B. 脱窒菌法 \end{CJK} \newline
\begin{CJK}{UTF8}{min} C. イオンクロマトグラフィー法 \end{CJK} \newline
\begin{CJK}{UTF8}{min} D. ガスクロマトグラフィー質量分析法 \end{CJK} & \begin{CJK}{UTF8}{min} B \end{CJK} \\\hline
\begin{CJK}{UTF8}{min} 遠隔転移は悪性腫瘍が全身化した状態で,治療の原則は薬物療法である。 \end{CJK} & \begin{CJK}{UTF8}{min} 遠隔転移は悪性腫瘍が全身化した状態で、治療の原則は[BLANK]である。 \end{CJK} & \begin{CJK}{UTF8}{min} 悪性腫瘍が全身化した状態である遠隔転移の治療の原則として用いられるのは何ですか？ \end{CJK} & \begin{CJK}{UTF8}{min} A. 放射線療法 \end{CJK} \newline
\begin{CJK}{UTF8}{min} B. 免疫療法 \end{CJK} \newline
\begin{CJK}{UTF8}{min} C. 外科的切除 \end{CJK} \newline
\begin{CJK}{UTF8}{min} D. 薬物療法 \end{CJK} & \begin{CJK}{UTF8}{min} D \end{CJK} \\\hline
\begin{CJK}{UTF8}{min} 門脈圧亢進症における消化管壁内粘膜下A-Vanastomosis (A-VA) 開大増加に伴う循環亢進状態に関しては, 食道胃静脈瘤・門脈圧亢進症性胃症の発症に直接関連する病態として, 多くの検討がなされてきている. \end{CJK} & \begin{CJK}{UTF8}{min} 門脈圧亢進症における消化管壁内粘膜下A-Vanastomosisの開大増加は[BLANK]に直接関連する病態として、多くの検討がなされてきている。 \end{CJK} & \begin{CJK}{UTF8}{min} 門脈圧亢進症における消化管壁内粘膜下A-Vanastomosisの開大増加が直接関連する病態として発症するのは何ですか？ \end{CJK} & \begin{CJK}{UTF8}{min} A. 肝性脳症および肝腎症候群の発症 \end{CJK} \newline
\begin{CJK}{UTF8}{min} B. 胆道閉塞および胆石症の発症 \end{CJK} \newline
\begin{CJK}{UTF8}{min} C. 膵炎および十二指腸潰瘍の発症 \end{CJK} \newline
\begin{CJK}{UTF8}{min} D. 食道胃静脈瘤および門脈圧亢進症性胃症の発症 \end{CJK} & \begin{CJK}{UTF8}{min} D \end{CJK} \\\hline
\begin{CJK}{UTF8}{min} BLT1を介したシグナルはMyD88の遺伝子発現を誘導することで腸内細菌からの自然免疫シグナルを増強し,形質細胞の細胞増殖を促進することで経口ワクチン抗原に対する抗原特異的IgA産生を促進する作用があることがわかり,経口ワクチンの成立に必須の分子機構であることが明らかになった. \end{CJK} & \begin{CJK}{UTF8}{min} BLT1を介したシグナルはMyD88の遺伝子発現を誘導することで[BLANK]を増強し、形質細胞の細胞増殖を促進することで経口ワクチン抗原に対する抗原特異的IgA産生を促進する作用があることがわかった。 \end{CJK} & \begin{CJK}{UTF8}{min} BLT1を介したシグナルがMyD88の遺伝子発現を誘導することで増強するのは何のシグナルですか？ \end{CJK} & \begin{CJK}{UTF8}{min} A. 腸管上皮細胞の増殖シグナル \end{CJK} \newline
\begin{CJK}{UTF8}{min} B. 抗原提示細胞による炎症性サイトカインシグナル \end{CJK} \newline
\begin{CJK}{UTF8}{min} C. 腸内細菌からの自然免疫シグナル \end{CJK} \newline
\begin{CJK}{UTF8}{min} D. ウイルス感染による獲得免疫シグナル \end{CJK} & \begin{CJK}{UTF8}{min} C \end{CJK} \\\hline
\hline
\end{tabularx}
\label{tab:adaxeval-ja-samples}
\end{table*}

\clearpage

\begin{table*}[htbp]
\scriptsize
\centering
\renewcommand{\arraystretch}{1.3} 
\caption{Samples of English AdaXEval Dataset.}
\begin{tabularx}{\textwidth}{X X X p{5cm} c}
\hline
\textbf{Sentence} & \textbf{cloze prompt} & \textbf{Paraphrase} & \textbf{Options} & \textbf{Answer ID} \\
\hline
Fe2O3-SiO2 particles, which removes 112S and COS in hot coal gas, are prepared. & Fe2O3-SiO2 particles, which remove [BLANK] in hot coal gas, are prepared. & Which contaminants in hot coal gas are targeted for removal by Fe2O3-SiO2 particles? & A. NH3 and HCN \newline
B. SO2 and NOx \newline
C. H2S and COS \newline
D. CO2 and CH4 & C \\\hline
The antimicrobial activity of lomefloxacin against gram-positive bacteria was inferior to those of ofloxacin and gentamicin and comparable to that of chloramphenicol. & The antimicrobial activity of lomefloxacin against gram-positive bacteria was inferior to those of [BLANK] and comparable to that of chloramphenicol. & Against gram-positive bacteria, lomefloxacin shows lower antimicrobial activity than which other antibiotics? & A. ciprofloxacin and ampicillin \newline
B. ofloxacin and gentamicin \newline
C. vancomycin and clindamycin \newline
D. erythromycin and tetracycline & B \\\hline
Basal cell adenoma is a rare type of salivary gland tumor. & [BLANK] is a rare type of salivary gland tumor. & Which rare type of tumor can occur in the salivary glands? & A. Adenoid cystic carcinoma \newline
B. Pleomorphic adenoma \newline
C. Basal cell adenoma \newline
D. Mucoepidermoid carcinoma & C \\\hline
The common clinical signs were fever and hepatosplenomegaly. & The common clinical signs were [BLANK]. & Which clinical signs are most frequently observed? & A. fever and hepatosplenomegaly \newline
B. jaundice and ascites \newline
C. rash and lymphadenopathy \newline
D. cough and chest pain & A \\\hline
Phospholipids were found to be classified into three groups : (1) a lipid deactivating the glycolipid by strong hydrogen bond (phosphatidic acid analog), (2) a lipid likely to distribute the glycolipid rather homogeneously by weak hydrogen bond (phosphatidylglycerol and phosphatidylinositol analogs), and (3) a lipid enhancing the activity of a glycolipid by electrostatic effect (phosphatidylserine, phosphatidylcholine, and phosphatidylethanolamine analogs). & Phospholipids were found to be classified into three groups: (1) a lipid deactivating the glycolipid by strong hydrogen bond (phosphatidic acid analog), (2) a lipid likely to distribute the glycolipid rather homogeneously by weak hydrogen bond (phosphatidylglycerol and phosphatidylinositol analogs), and (3) [BLANK]. & Which group of phospholipids enhances the activity of glycolipids by electrostatic effect? & A. lipids inhibiting glycolipid synthesis by covalent modification (ceramide analogs) \newline
B. lipids deactivating glycolipids by strong hydrogen bond (phosphatidic acid analog) \newline
C. lipids enhancing glycolipid activity by electrostatic effect (phosphatidylserine, phosphatidylcholine, and phosphatidylethanolamine analogs) \newline
D. lipids distributing glycolipids homogeneously by weak hydrogen bond (phosphatidylglycerol and phosphatidylinositol analogs) & C \\\hline
Furthermore, the dried ARC, which was dehydrated in the presence of saccharides, can be recovered by dispersion of the powdered ARC in water. & The dried ARC, which was dehydrated in the presence of saccharides, can be recovered by [BLANK]. & What process allows the recovery of dried ARC that was dehydrated with saccharides? & A. heating the powdered ARC in ethanol \newline
B. exposing the powdered ARC to ultraviolet light \newline
C. dispersion of the powdered ARC in water \newline
D. mixing the powdered ARC with organic solvents & C \\\hline
About half of the reported cases of acanthosis nigricans are accompanied by various kinds of malignant neoplasms, mostly adenocarcinomas of the digestive system. & About half of the reported cases of acanthosis nigricans are accompanied by [BLANK]. & What condition is present in about half of the reported cases of acanthosis nigricans? & A. various kinds of malignant neoplasms, mostly adenocarcinomas of the digestive system \newline
B. infectious diseases, such as tuberculosis and hepatitis \newline
C. autoimmune disorders, mainly lupus and rheumatoid arthritis \newline
D. benign skin tumors, primarily lipomas and fibromas & A \\\hline
Suppressed cellular immunity, malignancy, diabetes mellitus and history of antibiotic usage are significant predisposing factors for the development of esophageal candidiasis. & [BLANK] are significant predisposing factors for the development of esophageal candidiasis. & Which conditions are considered important risk factors for developing esophageal candidiasis? & A. Chronic hypertension, obesity, hyperlipidemia, and smoking \newline
B. Suppressed cellular immunity, malignancy, diabetes mellitus, and history of antibiotic usage \newline
C. Asthma, seasonal allergies, eczema, and vitamin D deficiency \newline
D. Alcohol abuse, liver cirrhosis, renal failure, and hypothyroidism & B \\\hline
Improved hydrogen sulfide removal is necessary for this apparatus to be applied to measurement of biogas produced by anaerobic digestion, since hydrogen sulfide influences catalysis. & Improved hydrogen sulfide removal is necessary for this apparatus to be applied to measurement of biogas produced by anaerobic digestion, since hydrogen sulfide influences [BLANK]. & What process is affected by the presence of hydrogen sulfide in biogas measurement apparatus? & A. oxidation \newline
B. fermentation \newline
C. catalysis \newline
D. photosynthesis & C \\\hline
The activity of ACC (1-aminocyclopropane-1-carboxylic acid) oxidase and the rate of ethylene production increased rapidly during fruit ripening at 20°C. & The activity of ACC (1-aminocyclopropane-1-carboxylic acid) oxidase and the rate of ethylene production increased rapidly during [BLANK]. & During which process at 20°C do ACC oxidase activity and ethylene production rate increase rapidly? & A. fruit ripening at 20°C \newline
B. leaf senescence at 20°C \newline
C. flowering at 20°C \newline
D. seed germination at 20°C & A \\\hline
\end{tabularx}
\label{tab:adaxeval-en-samples}
\end{table*}
\end{document}